\pdfoutput=1

\documentclass[11pt]{article}

\usepackage{ACL2023}

\usepackage{times}
\usepackage{latexsym}

\usepackage[T1]{fontenc}

\usepackage[utf8]{inputenc}

\usepackage{microtype}

\usepackage{inconsolata}

\usepackage{amsmath}
\usepackage{amsfonts}
\usepackage[ruled,vlined,linesnumbered]{algorithm2e}
\usepackage{booktabs}
\usepackage{tabularx}
\usepackage{xspace}
\usepackage{xcolor}
\usepackage{arydshln}

\usepackage{enumitem}
\usepackage{bbding}

\usepackage{graphicx}
\usepackage{subcaption}

\usepackage{appendix}
\usepackage{lipsum}

\usepackage{etoolbox}
\makeatletter
\patchcmd{\hyper@makecurrent}{%
    \ifx\Hy@param\Hy@chapterstring
        \let\Hy@param\Hy@chapapp
    \fi
}{%
    \iftoggle{inappendix}{
        \@checkappendixparam{chapter}%
        \@checkappendixparam{section}%
        \@checkappendixparam{subsection}%
        \@checkappendixparam{subsubsection}%
        \@checkappendixparam{paragraph}%
        \@checkappendixparam{subparagraph}%
    }{}%
}{}{\errmessage{failed to patch}}

\newcommand*{\@checkappendixparam}[1]{%
    \def\@checkappendixparamtmp{#1}%
    \ifx\Hy@param\@checkappendixparamtmp
        \let\Hy@param\Hy@appendixstring
    \fi
}
\makeatletter

\newtoggle{inappendix}
\togglefalse{inappendix}

\apptocmd{\appendix}{\toggletrue{inappendix}}{}{\errmessage{failed to patch}}
\apptocmd{\subappendices}{\toggletrue{inappendix}}{}{\errmessage{failed to patch}}

\newcommand\fl[1]{{\fontfamily{phv}\selectfont\footnotesize#1}}
\newcommand\eg[0]{\textit{e.g.}}
\newcommand\ie[0]{\textit{i.e.}}

\newcommand{\E}{$\mathcal{E\,}$}
\renewcommand{\L}{$\mathcal{L\,}$}
\newcommand{\R}{$\mathcal{R\,}$}
\newcommand{\C}{$\mathcal{C\,}$}

\newcommand{\OurMethod}{Pangu\xspace}

\newcommand{\Freebase}{{\textsc{Freebase}}\xspace}

\newcommand{\WebQ}{{\textsc{WebQ}}\xspace}
\newcommand{\WebQSP}{{\textsc{WebQSP}}\xspace}

\newcommand{\GraphQ}{{\textsc{GraphQ}}\xspace}
\newcommand{\GrailQ}{{\textsc{GrailQA}}\xspace}

\newenvironment{entityfont}{\fontfamily{qcr}\selectfont\footnotesize}{\par}
\newenvironment{relationfont}{\fontfamily{qcr}\selectfont\footnotesize}{\par}
\DeclareTextFontCommand{\textentity}{\entityfont}
\DeclareTextFontCommand{\textrelation}{\relationfont}
\newcommand\naturall[1]{{\it``#1''}} 

\newcommand{\iid}{i.i.d.\xspace}
\newcommand{\IID}{I.I.D.\xspace}

\newenvironment{remark}[1][Remark]{\begin{trivlist}
\item[\hskip \labelsep {\bfseries #1}]}{\end{trivlist}}

\newcommand{\nop}[1]{}

\title{\textit{Don't Generate, Discriminate}: \\A Proposal for Grounding Language Models to Real-World Environments}

\author{Yu Gu \\
  The Ohio State University \\
  \texttt{gu.826@osu.edu} \\\And
  Xiang Deng \\
  The Ohio State University \\
  \texttt{deng.595@osu.edu} \\\And
  Yu Su \\
  The Ohio State University \\
  \texttt{su.809@osu.edu} \\}

\begin{document}
\maketitle
\begin{abstract}
A key missing capacity of current language models (LMs) is grounding to real-world environments.
Most existing work for grounded language understanding uses LMs to directly generate plans that can be executed in the environment to achieve the desired effects. 
It thereby casts the burden of ensuring grammaticality, faithfulness, and controllability all on the LMs.
We propose \OurMethod, a generic framework for grounded language understanding that capitalizes on the discriminative ability of LMs instead of their generative ability.
\OurMethod consists of a symbolic agent and a neural LM working in a concerted fashion: The agent explores the environment to incrementally construct valid plans, and the LM evaluates the plausibility of the candidate plans to guide the search process.
A case study on the challenging problem of knowledge base question answering (KBQA), which features a massive environment, demonstrates the remarkable effectiveness and flexibility of \OurMethod: 
A BERT-base LM is sufficient for setting a new record on standard KBQA datasets, and larger LMs further bring substantial gains.
\OurMethod also enables, for the first time, effective few-shot in-context learning for KBQA with large LMs such as Codex. \footnote{Code and data will be released at \url{https://github.com/dki-lab/Pangu}.}

\end{abstract}

\section{Introduction}
Language models (LMs) such as BERT~\cite{devlin-etal-2019-bert}, GPT-3~\cite{brown2020language}, and Codex~\cite{chen2021codex} have demonstrated an extraordinary capacity in understanding and generating both natural language~\cite{Minaee2021DeepLB, liang2022holistic} and generic programs (\eg, Python)~\cite{Li2022CompetitionLevelCG, Jain2021JigsawLL, Austin2021ProgramSW}.
The recent release of ChatGPT is elevating this paradigm to a new level.\footnote{\url{chat.openai.com}}
It seems to point us towards a future where natural language serves as a universal device, powered by LMs, for automated problem solving and interacting with the (computing) world. 

\begin{figure}[t]
\centering
\includegraphics[width=\linewidth]{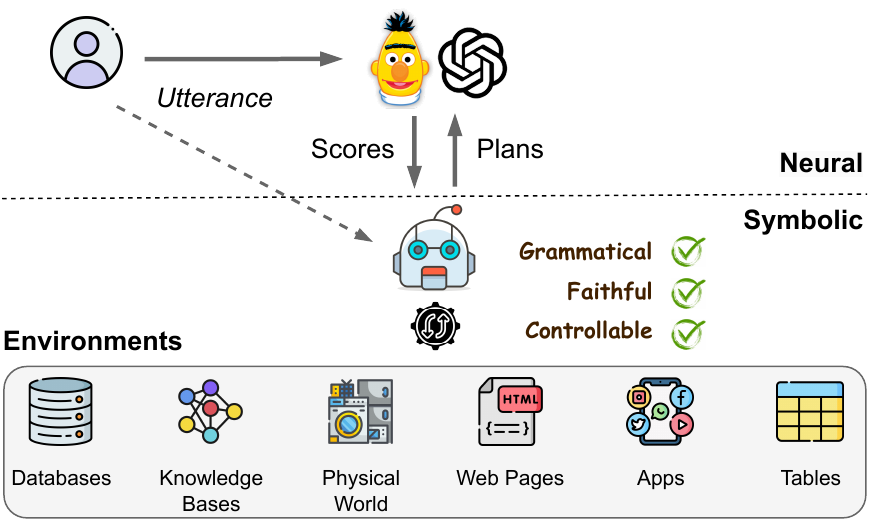}
\caption{A schematic illustration of the proposed framework, \OurMethod, where a symbolic agent interacts with the target environment to propose candidate plans, and a neural LM evaluates the plausibility of each plan. The agent searches the environment to incrementally construct the plans, and the LM guides the search process.}
\label{fig:overview}
\vspace{-10pt}
\end{figure}

However, a key missing piece in realizing this future is the connection between LMs and real-world environments, including both digital environments (\eg, databases, knowledge bases, Excel spreadsheets, software, websites, among others) and physical environments (\eg, instruction following robots \cite{shridhar2020alfred,ahn2022can}).
Such environments are where many real problems lie.
For example, 
a biologist may need to find all the species of a certain butterfly genus and their geographic distribution from a biology knowledge base,
a local grocery store owner may want to visualize the historical sales of different item categories in Excel to decide what and how much to restock before the holiday season, and a physician may need to find patients with specific conditions in a large database of electronic medical records 
to inform the current diagnosis.
\textit{How can LMs enable solving all these problems, which involve seeking information or taking actions in a specific environment, with natural language?}

Each environment is a unique context for interpreting natural language requests from users. 
\textit{Grounding}, \ie, linking of (natural language) concepts to contexts~\cite{chandu-etal-2021-grounding}, therefore becomes the fundamental problem.
More precisely, we need to produce a \textit{plan} (also called a \textit{program} when described using a programming language) that can be executed in an environment to achieve the desired effects of the corresponding language request. 
The unique challenge of such \textit{grounded language understanding} problems stems from 1) the vast heterogeneity of environments and their planning languages (\eg, SQL, GraphQL/REST APIs, $\lambda$-calculus, and robot planning languages), and 2) the vast, oftentimes infinite, number of possible instantiations (or states) of each environment.
Some environments can also be dynamic (\eg, a database that is constantly updated or a physical environment with moving objects).

Most existing methods for grounded language understanding follow the popular sequence-to-sequence framework~\cite{seq2seq,cho-etal-2014-learning} and generate the plans/programs in an autoregressive fashion~\cite{ xie2022unifiedskg, ye-etal-2022-rng, wang-etal-2021-codet5, song2022one}.
A core thesis of this paper is that \textit{directly generating plans may not be the optimal way of using LMs for grounded language understanding}. 
It requires LMs to have intimate knowledge about each specific planning language and  environment, neither of which may be part of an LM's pre-training, to ensure the \textit{grammaticality} (\ie, conforming to the grammar of the planning language) and \textit{faithfulness} (\ie, executable in the environment) of the generated plans. 
The infinite and dynamic environment states also reduce the potential effectiveness of pre-training for improving faithfulness, even if one manages to do so. 
Furthermore, autoregressive generation with a neural LM lacks fine-grained \textit{control} over planning; it is cumbersome, though not impossible, to factor preferences, business logic, and other values and constraints into the plan generation process. 
A focus of recent work is to alleviate (some of) these limitations by augmenting autoregressive generation with environment-specific pre-training~\cite{yu2021grappa,deng-etal-2021-structure} or constrained decoding~\cite{scholak-etal-2021-picard, shin-etal-2021-constrained,gu-su-2022-arcaneqa}.
However, the fundamental challenges still largely remain.

Mathematically, an LM is simply a joint distribution $p(x_1, x_2, ..., x_n)$ that factors as a product of conditional distributions $\prod_{i=1}^{n} p(x_i | x_1, ..., x_{i-1})$. 
Existing work leverages the conditional distribution formulation to generate the plan.
It thereby casts the burden of ensuring grammaticality, faithfulness, and controllability all on the LM. 
The main proposal of this paper is to \textit{disentangle LMs from these responsibilities and let LMs be what they originally are---a model that assigns a probability to a sequence of tokens}. 
In other words, we advocate for using the joint distribution formulation of LMs to evaluate the plausibility of (utterance, candidate plan) pairs instead of directly generating the plan. 

To this end, we propose \OurMethod, a generic framework for grounded language understanding that capitalizes on the discriminative ability of LMs instead of their generative ability (\autoref{fig:overview}).\footnote{Pangu is a primordial being in Chinese mythology who separated heaven and earth. We name our framework after that for its separating the realm of the neural and the symbolic.}
\OurMethod consists of a symbolic agent and a neural LM working in a concerted way. 
The symbolic agent explores the environment to propose candidate plans, which are guaranteed by design to be both grammatical and faithful. 
For most real-world environments, due to the size of the search space or partial observability, it is necessary for the agent to search in the environment and incrementally extend or refine the plans.
The LM plays a key role in this search process---it evaluates the candidate (partial) plans at each search step and guides the agent towards promising search directions; it also determines when the search ends.
Finally, it is also easier to control the search process of a symbolic agent than the generation process of a neural LM. 

As a case study, we instantiate the proposed framework for complex question answering over knowledge bases (KBQA). 
KBQA provides an ideal testbed for grounded language understanding because of its massive environment---direct generation with LMs often fails dramatically~\cite{gu2021beyond}. 
We show that simply using BERT-base with \OurMethod is sufficient for setting a new record on standard KBQA datasets, and larger LMs further bring substantial gains. 
\OurMethod also enables, for the first time, few-shot KBQA by prompting large language models (\eg, Codex): 
Using only 10 labeled examples, it outperforms all prior methods on \GraphQ~\cite{su-etal-2016-generating}.
It provides \textit{unprecedented uniformity} for using LMs---one can easily plug encoder-only LMs, encoder-decoder LMs, or decoder-only LMs into \OurMethod. 
These results highlight the remarkable effectiveness and flexibility of \OurMethod and validate the proposal of using LMs for discrimination instead of generation.

\section{Related Work}
\noindent \textbf{Generation for Grounded Language Understanding.} The Seq2Seq framework~\cite{seq2seq,bahdanau2014neural} has been the \textit{de facto} choice for grounded language understanding, where the LM directly generates a plan given an input utterance. 
However, the lack of grounding during pre-training makes generating valid plans from the LM challenging. Recent studies endeavor to alleviate this issue via \emph{input augmentation} or \emph{constrained decoding}. 
For input augmentation, the environment (or some relevant portion of it) is fed to the LM's encoder together with the utterance~\cite{hwang2019comprehensive,wang-etal-2020-rat,xie2022unifiedskg}. Such methods rely on the LM to understand the interplay between language requests and the environment and correctly factor that into plan generation. 
They therefore require substantial training data to learn and also provide no guarantee for grammaticality or faithfulness. 
In contrast, constrained decoding methods regulate the decoder's behavior to guarantee grammaticality~\cite{scholak-etal-2021-picard,shu2022tiara} or even faithfulness~\cite{liang-etal-2017-neural,gu-su-2022-arcaneqa}. 
However, such uses still cast the burden of generating valid plans on the LM itself; controlling the generation process of an LM can be difficult and specific to each planning language and/or environment. 
In our proposal, the LM is only used to discriminate valid plans proposed by an agent through a controllable search process. 
More comparison is discussed in \S\ref{sec:constrained_decoding}.

\nop{Different from inference-time augmentation, pre-training augmentation methods seek to reduce the gap between LMs' pre-training settings and grounded language understanding tasks by performing environment-specific pre-training~\cite{yu2021grappa,deng-etal-2021-structure}. However, these methods need a large set of environment-specific data for pre-training (\eg, aligned corpus for table schema and utterances~\cite{yu2021grappa}) and do not generalize to different environments (\eg, LMs with table-specific pre-training are not applicable to knowledge bases). 
By contrast, \OurMethod requires no pre-training and can be easily plugged into different environments.}

\nop{The use of Large Language Models (LLMs) such as GPT-3~\cite{brown2020language} and Codex~\cite{chen2021codex} in grounded language understanding tasks has attracted increasing interest recently.} 

\noindent \textbf{Few-Shot Grounded Language Understanding with LLMs.}
Large language models (LLMs)~\cite{brown2020language,chen2021codex} have demonstrated strong few-shot learning capabilities in various tasks, from writing programs to query structured and unstructured data~\cite{Austin2021ProgramSW, Rajkumar2022EvaluatingTT, cheng2022binding}, interacting with \nop{mobile UI~\cite{Wang2022EnablingCI} and} online websites~\cite{Gur2022UnderstandingHW, Nakano2021WebGPTBQ}, to generating procedural plans and guiding embodied agents in virtual environments~\cite{singh2022progprompt, ahn2022can, shah2022lmnav, song2022llm}. 
Most existing work still capitalizes on the generative ability of LLMs. 
A common strategy to encourage an LLM to produce valid plans is to directly \textit{describe} the environment in the LLM's context (\ie, input augmentation), which is difficult for complex environments like KBs. 
In contrast, \OurMethod shields the LLM from the complexity of the environment and lets the LLM focus on evaluating the plausibility of candidate plans proposed by an agent. 
One interesting related work is \citet{ahn2022can}, where an LLM is used to score atomic action (skill) proposals, which are guaranteed to conform to affordance constraints, from an embodied agent. \OurMethod shares a similar spirit of using LMs for discrimination, but we support more complex plans through a search process in the environment guided by an LM.

\noindent \textbf{Bottom-Up Semantic Parsing.}
Our instantiation of \OurMethod on KBQA is closely connected to bottom-up semantic parsing, particularly SmBoP~\cite{rubin-berant-2021-smbop}, a text-to-SQL model that iteratively constructs a complex plan from a set of subplans. \nop{At each step of parsing, SmBoP enumerates candidate parse trees from all valid combinations of trees from the previous step and scores them to get the top-ranked ones.}\OurMethod similarly constructs a complex plan incrementally from smaller subplans, but it makes the following main departures. First, SmBoP requires all ingredients (\ie, column headers, table names, and DB values) at the beginning of parsing. This assumption does not generally hold for more complex or partially observable environments, where ingredients need to be discovered through search. 
In our method, only topic entities are needed as the initial plan, which can be readily obtained using an entity linker~\cite{li-etal-2020-efficient}. 
Second, our scoring function is based on a straightforward application of LMs, while SmBoP uses a more intricate architecture with extra parameters. 
Also related is an array of earlier KBQA methods that adopt an enumerate-and-rank approach~\cite{yih-etal-2015-semantic,gu2021beyond,ye-etal-2022-rng}.
Because they try to enumerate all candidate plans up front, the maximum plan complexity is bound to be small. 
Our adaptive search process allows for flexible construction of more complex plans.

\section{Approach}
\begin{figure*}[!t]
\centering
\includegraphics[width=0.95\linewidth]{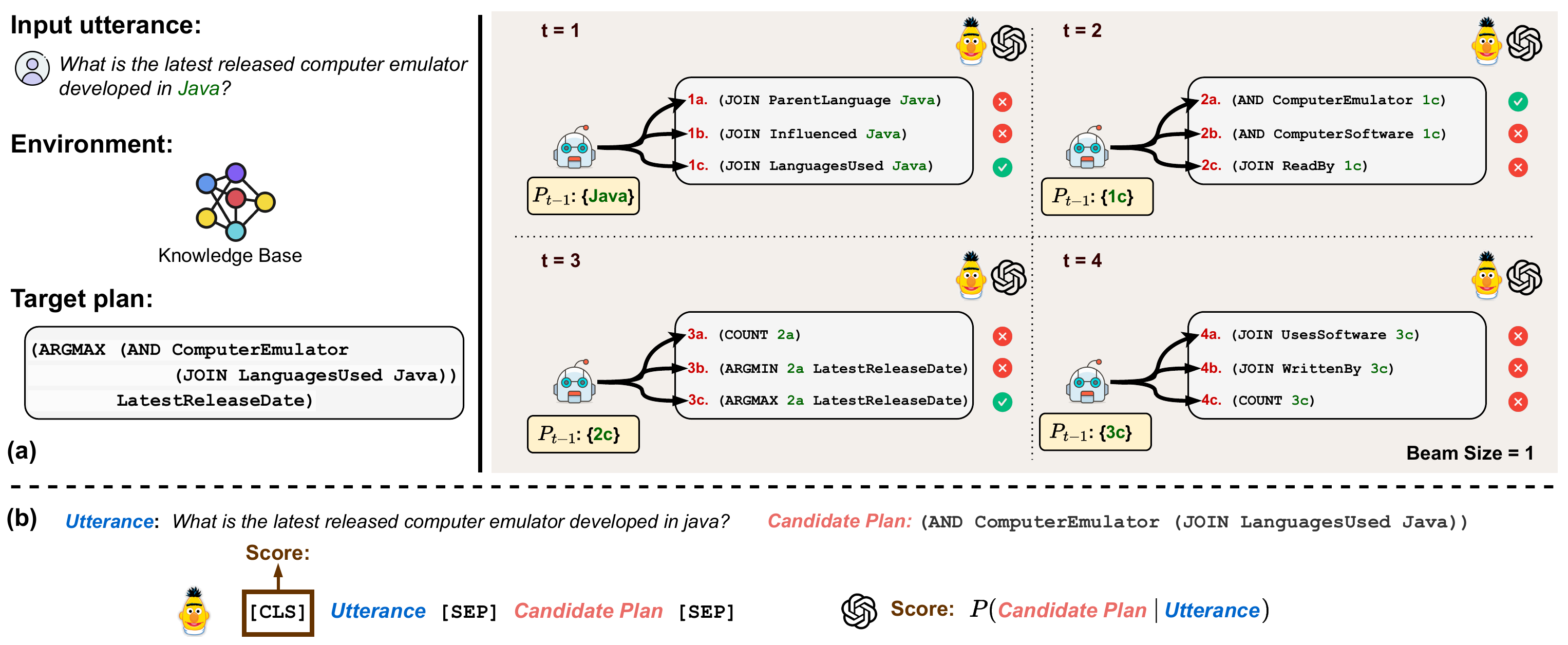}
\vspace{-1em}
\caption{(a) An illustration of how an agent collaborates with an LM to incrementally produce a complex target plan over a KB using beam search (beam size = 1 in this example). At each step, the agent enumerates a set of valid plans based on the current plans and the environment. An LM then scores the candidate plans and returns the top-ranked ones. The search process terminates when there is no candidate plan that scores higher than the current best plan (\eg, 4a-c are all worse than 3c). (b) Using different LMs (\textit{left}: BERT, \textit{right}: Codex) to evaluate the plausibility of plan 2a. It resembles using LMs for semantic matching between the utterance and the plan.}
\vspace{-1em}
\label{fig:loop}
\end{figure*}

An overview of the \OurMethod framework is presented in~\autoref{alg:highlevel}. 
An overarching assumption of \OurMethod is that a complex plan can be incrementally constructed by an agent through its exploration in an environment. 
Such an agent can be a robot doing household tasks in a physical environment~\cite{shridhar2020alfred}, or a virtual agent that orchestrates API calls of different web services~\cite{andreas-etal-2020-task} or traverses a database/knowledge base (KB)~\cite{yu-etal-2018-spider,gu2022knowledge}. 
Starting from a set of initial plans $P_0$ (may be empty), at each step, the agent interacts with the environment $E$ to extend the current plans into a new set of candidate plans (\autoref{line:enumerate}). 
The candidate plans are guaranteed to be \textit{valid} (\ie, both grammatical and faithful).
An LM then scores the candidate plans, and the top $K$ (the beam size) plans are retained for further exploration in the next step (\autoref{line:matching}). 
The same procedure loops until a termination check is passed (\autoref{line:stop}); the best plan is then returned. 

\OurMethod mainly shines in that a symbolic agent explores the environment to propose valid plans and shields the LM from having to handle the large search space for valid plan generation. 
Instead, the LM only focuses on evaluating the plausibility of the proposed plans\nop{, \ie, to what extent a candidate plan matches the intent of the input utterance}. 
An LM can be easily fine-tuned to excel at this assignment, or, in the case of LLMs such as Codex, they come with such ability out of the box, which enables few-shot in-context learning. 
\OurMethod is a generic framework and can potentially accommodate many grounded language understanding tasks by instantiating the various functions in~\autoref{alg:highlevel} accordingly.

\subsection{KBQA: Preliminaries}
Without loss of generality, we use KBs as our target environment and the knowledge base question answering (KBQA) task as a concrete example for ease of discussion (we discuss possible implementation for other tasks in~\autoref{appendix:others}). 
It is an ideal testbed because of the massive environment provided by modern KBs (\eg, \Freebase~\cite{bollacker2008freebase} contains 45 million entities and 3 billion facts for over 100 domains), which makes grounding particularly challenging. 
Given a KB $\mathcal{K}\subset\mathcal{E}\times\mathcal{R}\times(\mathcal{E}\cup\mathcal{L}\cup\mathcal{C})$, where \C is a set of classes, \E a set of entities, \L a set of literals and \R a set of binary relations, the task of KBQA is to find a set of answer entities to an input utterance in the KB. 
KBQA is typically modeled as semantic parsing~\cite{gu2022knowledge}, where the utterance is mapped to an executable program/plan in a certain formal language (e.g., SPARQL, $\lambda$-calculus, or S-expression) whose denotation is the answer.\nop{\footnote{we will use plan and program interchangeably hereafter.}}
We use S-expressions~\cite{gu2021beyond} for its compactness.
An example is shown in~\autoref{fig:loop}.
\begin{algorithm}[!t]
\small
\textbf{Input:} utterance $q$, initial plans $P_0$, environment $E$  \\
    $t \leftarrow 1$; \\
    \While{True}{
        \tcp{\textsc{Agent proposes plans}}
        $C_{t} \leftarrow
        \textbf{Candidate-Plans}(P_{t-1}, E)$  \label{line:enumerate}\\
        \tcp{\textsc{LM scores and prunes plans}}
        $P_{t} \leftarrow \textbf{Top-}K(q, C_{t})$\label{line:matching}\\
        \If{\textbf{Check-Termination}() = True \label{line:stop}}{\Return top-scored plan}
        $t \leftarrow t + 1$ \\
 }
\caption{\textsc{\OurMethod}}
\label{alg:highlevel}
\end{algorithm}
\setlength{\textfloatsep}{0.1pt}

\subsection{Candidate Plan Enumeration}
To handle the large search space, the agent casts the task as a step-wise decision-making problem. \nop{For example, an instruction-following robot may decompose a plan into a sequence of subplans (\eg, making a cup of coffee entails first finding a cup then picking up the cup, etc.; \citet{song2022llm}).}
A plan for KBQA can be decomposed into a nested sequence of subplans~\cite{gu-su-2022-arcaneqa} (\autoref{fig:loop}). 
The \emph{length} of a plan is defined as the number of atomic subplans it contains.

For KBQA, $P_0$ can be a set of entity proposals (\eg, \{\fl{Java}\}) obtained using off-the-shelf entity linkers~\cite{li-etal-2020-efficient}.
At step $t$, the agent considers $P_{t-1}$, the length $t-1$ plans, and decides how to further extend them into $C_{t}$, the valid plans of length $t$, based on the environment. 
This often involves executing the current plans in the environment.
Consider the example in \autoref{fig:loop} at $t=1$, the agent finds all the relations connected to \fl{Java} and enumerates all the length-1 valid plans.
The LM scores the candidate plans and prunes all but the top-ranked plan because beam size is 1. 
At $t=2$, the agent executes plan $1c$ to get its denotation (\ie, a set of entities) in the KB, based on which the agent further discovers the relations and classes (\eg, \fl{ComputerEmulator}, \fl{ComputerSoftware}, and \fl{ReadBy}) connected to those entities to form valid length-2 plans.
All the plans produced in this process are guaranteed to be valid.
See~\autoref{appendix:candidate} for a more detailed discussion of this process.

\subsection{LM-Based Scoring}\label{section:lm-evaluation}
After the agent enumerates a set of candidate plans, an LM assists with its decision making by evaluating the plausibility of each candidate plan. The interface for evaluating a plan using LMs resembles using LMs for semantic matching: Given a pair of ($u\colon$\textit{utterance}, $c\in C_t\colon$\textit{candidate plan}), an LM acts as a scoring function: $s(u, c)\rightarrow\mathbb{R}$, which indicates to what extent the candidate plan matches the intent of the utterance.  
The plausibility of a candidate oftentimes can be indicated by simple linguistic cues, \eg, \fl{ComputerEmulator} in $2a$ might be a strong indicator (\autoref{fig:loop}(a)).

We follow the common practice of using LMs for semantic matching. For encoder-only LMs like BERT, we directly get a score from the representation of the \fl{[CLS]} token (\autoref{fig:loop}(b)). 
For encoder-decoder LMs like T5, we follow~\citet{zhuang2022rankt5} to feed both the utterance and the candidate plan to the encoder and use the decoding probability over an unused token\nop{\footnote{We use \textrelation{<extra\_id\_23>} for T5.}} during pre-training as a proxy for matching score. For decoder-only LMs like Codex, we model the score as the probability of generating the candidate plan conditioned on the utterance, \ie, $P(c|u)$. 
Intuitively, a good scoring function should respect the following \textit{partial order}:
\vspace{-15pt}
\begin{align}
    & s(u, c_1) > s(u, c_2),\quad \forall c_1\in G_t \textrm{ and } \forall c_2\in G_{t-1}, \nonumber \\
    & s(u, c_1) > s(u, c_2),\quad \forall c_1\in G_t \textrm{ and } \forall c_2\in C_t\backslash G_t, \nonumber \\
    & s(u, c^\prime) > s(u, c_i), \hspace{0.12cm} \quad \forall c_i \neq c^\prime \nonumber
\end{align}
where $G_t$ is the set of gold (sub-)plans at step $t$ (\ie, length-$t$ subplans of the target plan) and $c^\prime$ is the target plan. In other words, a gold subplan should be scored higher than 1) any negative (\ie, not gold) plans at the same step (\eg, $2a$ should be scored higher than $2c$), because they contain information irrelevant to $u$, and 2) any gold sub-plans of length $<t$ (\eg, $2a$ should be scored higher than $1c$) because they are less complete. In addition,  $c^\prime$ should be scored higher than any other plan.

\subsection{Termination Check}
Assuming the LM can assign reasonable scores to candidate plans following the above partial order, we can naturally define the condition for termination in~\autoref{alg:highlevel}: It terminates if the highest score of candidate plans at step $t$ is lower than the highest score of candidate plans at step $t-1$, which, ideally, should indicate no reachable candidate plan of length $\geq t$ is better than the plans at step $t-1$, and thus the search process terminates.
\nop{It is worth noting that this may not be the only way of checking termination for other grounded language understanding tasks.
For example, an instruction-following robot may check the environment state (\eg, whether a cup of coffee has been successfully made) for that purpose.}

\subsection{Learning}
We discuss the learning procedure for both fine-tuning LMs (\eg, T5) and in-context learning with LLMs (\eg, Codex). 
For both settings, we use pairs of utterances and gold plans for supervision.

\begin{remark}[Fine-tuning.]
Given a gold plan of length $T$, we first derive its gold sub-plans $G_t$ of each step $t\leq T$ (\eg, $1c$ for step 1 and $2a$ for step 2 in~\autoref{fig:loop}). 
Fine-tuning proceeds with beam search similar to the test-time behavior, but with bottom-up teacher forcing~\cite{williams1989learning,rubin-berant-2021-smbop}, \ie, the gold plans of the current step should always be inserted into the beam.  
At each step of beam search, we get the probability of each candidate plan $c\in C_t$ with softmax over the scores: $p(c) = \text{softmax}\{s(u, c)\}_{c\in C_t\cup G_{t-1}}$. $G_{t-1}$ is also included here to encourage LMs to explicitly learn the partial order by minimizing the loss:
\begin{equation}
    -\frac{1}{Z}\sum_{t=1}^{T+1}\sum_{c\in C_t}\hat{p}(c)\text{log }p(c)\nonumber
\end{equation}
where $Z$ is the total number of summed items, and $\hat{p}(c)$ equals 1 if $c\in G_t$ and 0 elsewise. Note that, for the $T+1$ step, we let $G_{T+1}=G_T$. This additional step aims to enforce the third condition in the partial order. Our objective is essentially a listwise learning-to-rank objective based on the cross entropy~\cite{cao2007learning}. 
\end{remark}

\begin{remark}[In-Context Learning.]
We directly use pairs of utterances and gold plans as in-context demonstrations to the LLM, with a simple task instruction in the prompt: \naturall{Please translate the following questions to Lisp-like programs.} 
The LLM is therefore expected to capture the desired partial order by observing the in-context examples.
For concrete examples of prompts, please refer to~\autoref{appendix:prompts}.
\end{remark}

\begin{table*}[!th]
\small
\begin{subtable}{\linewidth}
\centering
\resizebox{0.92\textwidth}{!}{\begin{tabular*}{\textwidth}{l@{\extracolsep{\fill}}cccccccccc}
\toprule
& \multicolumn{2}{c}{\textbf{Overall}}   &\multicolumn{2}{c}{\textbf{\IID}}  &\multicolumn{2}{c}{\textbf{Compositional}}    &\multicolumn{2}{c}{\textbf{Zero-shot}}  &\multicolumn{2}{c}{\textbf{Dev Overall}}     \\ \cmidrule{2-11}
\multicolumn{1}{l}{\textbf{Model}} & \textbf{EM} & \textbf{F1} & \textbf{EM} & \textbf{F1} & \textbf{EM} & \textbf{F1} & \textbf{EM} & \textbf{F1} & \textbf{EM}  &\textbf{F1}\\ 
\midrule
\multicolumn{1}{l}{QGG~\cite{lan-jiang-2020-query}} & $-$    & 36.7  & $-$ & 40.5  & $-$   & 33.0  & $-$ &  36.6  & $-$ & $-$  \\
\multicolumn{1}{l}{BERT+Ranking~\cite{gu2021beyond}}      & 50.6  &  58.0     & 59.9  & 67.0       & 45.5  &   53.9    & 48.6  &  55.7 & $-$ & $-$  \\
\multicolumn{1}{l}{ReTraCk~\cite{chen-etal-2021-retrack}}      & 58.1  &  65.3     & 84.4  & 87.5       & 61.5  &   70.9    & 44.6  &  52.5 & $-$ & $-$  \\
\multicolumn{1}{l}{RnG-KBQA~\cite{ye-etal-2022-rng}}      & 68.8  &  74.4     & 86.2  & 89.0       & 63.8  &   71.2    & 63.0  &  69.2 & 71.4 & 76.8  \\
\multicolumn{1}{l}{ArcaneQA~\cite{gu-su-2022-arcaneqa} } &63.8 &73.7 & 85.6 & 88.9 & 65.8 & 75.3 & 52.9 &66.0 & 69.5 & 76.9\\
\multicolumn{1}{l}{Uni-Parser~\cite{liu2022uni}} & 69.5 & 74.6 & 85.5 & 88.5 & 65.1 & 71.1 & 64.0 & 69.8 & 70.8 & 76.5\\
\multicolumn{1}{l}{TIARA~\cite{shu2022tiara}} &73.0 &78.5 & \textbf{87.8} & \textbf{90.6} & 69.2 & 76.5 & 68.0 &73.9 & 75.3 & 81.9\\
\multicolumn{1}{l}{DecAF~\cite{yu2022decaf}} &68.4 &78.7 &84.8  & 89.9 &73.4  & 81.8 &58.6  & 72.3 & $-$ & 81.4\\
\multicolumn{1}{l}{UnifiedSKG w/ T5-3B~\cite{xie2022unifiedskg}} 
&$-$ & $-$ & $-$ & $-$ & $-$ & $-$ & $-$ & $-$ & 70.1$^*$ & $-$\\
\midrule

\OurMethod (this work) & & & & & & & & \\
\multicolumn{1}{l}{\hspace{10pt} w/ BERT-base} &73.7 & 79.9 & 82.6 & 87.1 & 74.9 & 81.2 & 69.1 & 76.1 & 75.0 & 82.1\\
\multicolumn{1}{l}{\hspace{10pt} w/  T5-base} &73.6 & 79.9 & 84.7 & 88.8 & 73.1 & 80.1 & 68.6 & 75.8 & \textbf{76.0} & 82.8\\
\multicolumn{1}{l}{\hspace{10pt} w/  T5-large} &74.8 & 81.4 & 82.5 & 87.3 & \textbf{75.2} & \textbf{82.2} & 71.0 & 78.4 & 75.8 & 83.3\\
\multicolumn{1}{l}{\hspace{10pt} w/  T5-3B} &\textbf{75.4} & \textbf{81.7} & 84.4 & 88.8 & 74.6 & 81.5 & \textbf{71.6} & \textbf{78.5} & 75.8 & \textbf{83.4}\\
\hdashline
\multicolumn{1}{l}{\hspace{10pt} w/  Codex (10-shot)}
& 48.9 & 56.3 & 51.8 & 58.1 & 43.3 & 51.2 & 50.1 & 57.8 & $-$ & $-$ \\
\multicolumn{1}{l}{\hspace{10pt} w/  Codex (100-shot)}
&53.3 & 62.7 & 54.7 & 62.9 & 54.5 & 63.7 & 52.3 & 62.2 & $-$ & $-$ \\
\multicolumn{1}{l}{\hspace{10pt} w/  Codex (1000-shot)}
&56.4 & 65.0 &  67.5 &  73.7 & 58.2 & 64.9 & 50.7 & 61.1  & $-$ & $-$ \\

\bottomrule
\end{tabular*}}
\caption{\GrailQ}
\label{table:overall_grail}
\end{subtable}
\begin{subtable}{0.47\linewidth}
\centering
 \resizebox{0.75\textwidth}{!}{\begin{tabular}{lc}
    \toprule
      \multicolumn{1}{l}{\textbf{Model}}  & \textbf{F1} \\
    \midrule
        UDEPLAMBDA~\cite{reddy-etal-2017-universal} & 17.7$^\sharp$ \\
        PARA4QA~\cite{dong-etal-2017-learning} & 20.4$^\sharp$ \\
        SPARQA~\cite{sun2020sparqa} & 21.5$^\sharp$ \\
        BERT+Ranking~\cite{gu2021beyond} & 27.0 \\
        ArcaneQA~\cite{gu-su-2022-arcaneqa}  & 34.3 \\
    \midrule
        \OurMethod (this work)  &  \\
\multicolumn{1}{l}{\hspace{10pt} w/ BERT-base} & 52.0\\
\multicolumn{1}{l}{\hspace{10pt} w/ T5-base} &  53.3\\
\multicolumn{1}{l}{\hspace{10pt} w/ T5-large} & 55.6\\
\multicolumn{1}{l}{\hspace{10pt} w/ T5-3B} & \textbf{62.2}\\
\hdashline
\multicolumn{1}{l}{\hspace{10pt} w/ Codex (10-shot)} & 42.8 \\
\multicolumn{1}{l}{\hspace{10pt} w/ Codex (100-shot)} & 43.3\\
\multicolumn{1}{l}{\hspace{10pt} w/ Codex (1000-shot)} & 44.3\\
    \bottomrule
    \end{tabular}}
    \caption{\GraphQ}
\label{table:overall_gq}
\end{subtable}
\begin{subtable}{0.6\linewidth}
\centering
 \resizebox{.55\textwidth}{!}{\begin{tabular}{lc}
    \toprule
       \multicolumn{1}{l}{\textbf{Model}} & \textbf{F1} \\
    \midrule
        QGG~\cite{lan-jiang-2020-query} & 74.0 \\
        ReTraCk~\cite{chen-etal-2021-retrack} & 71.0 \\
        CBR~\cite{das-etal-2021-case} & 72.8 \\
        Program Transfer~\cite{cao-etal-2022-program} & 76.5$^*$ \\
        RnG-KBQA~\cite{ye-etal-2022-rng} & 75.6  \\
        ArcaneQA~\cite{gu-su-2022-arcaneqa} & 75.6 \\
        Uni-Parser~\cite{liu2022uni} & 75.8 \\
        TIARA~\cite{shu2022tiara}  & 76.7 \\
        DecAF~\cite{yu2022decaf} & 78.8 \\
    \midrule
       \OurMethod (this work)  &  \\
\multicolumn{1}{l}{\hspace{10pt} w/ BERT-base} & 77.9\\
\multicolumn{1}{l}{\hspace{10pt} w/ T5-base} & 77.3\\
\multicolumn{1}{l}{\hspace{10pt} w/ T5-large} & 78.9\\
\multicolumn{1}{l}{\hspace{10pt} w/ T5-3B} & \textbf{79.6}\\
\hdashline
\multicolumn{1}{l}{\hspace{10pt} w/ Codex (10-shot)} &  45.9 \\
\multicolumn{1}{l}{\hspace{10pt} w/ Codex (100-shot)} & 54.5  \\
\multicolumn{1}{l}{\hspace{10pt} w/ Codex (1000-shot)} & 68.3 \\
    \bottomrule
    \end{tabular}}
        \caption{\WebQSP}
\label{table:overall_wq}
\end{subtable}
\vspace{-1em}
\caption{Overall results. \OurMethod achieves a new state of the art on all three datasets and shows great flexibility in accommodating LMs of different nature. Also, for the first time, \OurMethod enables effective few-shot in-context learning for KBQA with Codex. $^*$ using oracle entity linking. $^\sharp$ results on the original \GraphQ 2013-07, otherwise it uses the version from \citet{gu-su-2022-arcaneqa}, which is a slightly smaller subset.}
\label{table:overall}
\vspace{-1em}
\end{table*}

\section{Experimental Setup}

\subsection{Datasets}
We experiment with three KBQA datasets of different scale and nature (statistics in \autoref{table:datasets}).

\noindent \textbf{\GrailQ}~\cite{gu2021beyond} is a large-scale dataset that evaluates three levels of generalization, namely, \textit{\iid}, \textit{compositional} (novel compositions of seen constructs), and \textit{zero-shot} (totally novel domains). 
It also features diverse questions of different complexity \nop{(\eg, programs may involve up to 4 relations)} and aggregation functions\nop{ (\eg, comparatives, superlatives, and counting)}. 

\noindent \textbf{\GraphQ}~\cite{su-etal-2016-generating} is a moderate-scale dataset. Due to the small size of its training set and the non-\iid\ setting, \GraphQ is particularly challenging. In our experiments, we use the processed version by~\citet{gu-su-2022-arcaneqa}, which maps the original dataset from \Freebase 2013-07 to \Freebase 2015-08-09. 

\noindent \textbf{\WebQSP}~\cite{yih-etal-2016-value} is a moderate-scale dataset with questions from Google query logs. 
It mainly tests \iid\ generalization on simple questions. It is a clean subset of \WebQ\cite{berant-etal-2013-semantic} with program annotations. 

\subsection{Baselines}
We mainly compare \OurMethod with state-of-the-art baselines that use LMs as a generative model, including ArcaneQA~\cite{gu-su-2022-arcaneqa}, TIARA~\cite{shu2022tiara}, DecAF~\cite{yu2022decaf}, and RnG-KBQA~\cite{ye-etal-2022-rng}. Constrained decoding (\ie, ArcaneQA and TIARA) and input augmentation (\ie, TIARA, DecAF) are used to enhance plan generation. Also, the last three models use a combination of language models to do different jobs (\ie, retrieval/ranking/decoding).
In addition, we also compare with UnifiedSKG~\cite{xie2022unifiedskg}. UnifiedSKG assumes a set of schema items are provided as input, where the gold schema items are always included and the number of negative schema items is restricted to 20 for \GrailQ. 
It is thus a less fair comparison for other methods, but we include it anyway because it is a representative way of autoregressive plan generation using a large LM.
\nop{There is no constrained decoding for the last three baseline methods, and thus neither grammaticality nor faithfulness can be guaranteed. 
These methods rely on labeled training data to learn how to produce valid programs; their sample efficiency suffers as a result (\S\ref{sec:sample_efficiency}).}
Compared with the baselines, \OurMethod requires no extra parameter, no modification to the LM, and no need to combine multiple LMs. \OurMethod provides unprecedented uniformity of using LMs of different nature.
More details on baselines can be found in~\autoref{appendix:baselines}.

\subsection{Implementation Details}
For the fine-tuning experiments, we experiment with BERT-base, T5-base, T5-large, and T5-3B, and use the full training set of each dataset for fine-tuning. For the in-context learning experiments, we experiment with Codex.\footnote{We opt for Codex because it is free, but small-scale experiments also show competitive performance from GPT-3.} 
We randomly sample 10/100/1000 training examples from each dataset and use that as the pool for dynamic retrieval\nop{, therefore a \textit{100-shot} setting}. 
During inference, for each test example, we retrieve 10 in-context examples from the pool using BM25-based utterance similarity. 
We use entity linking results from off-the-shelf entity linkers. More details on implementations can be found in~\autoref{appendix:implementation}.

\section{Results}
\subsection{Main Results}
\begin{remark}[Fine-tuning results.]
The main results are shown in~\autoref{table:overall}. 
Using a BERT-base LM, \OurMethod already achieves a new state of the art on \GrailQ and \GraphQ, and only trails behind DecAF on \WebQSP, which uses a 3B-parameter LM. 
On \GraphQ, \OurMethod with BERT-base dramatically improves the state-of-the-art F1 from 31.8\% to 48.2\%. 
These are strong evidence for \OurMethod being a better protocol for using LMs for grounded language understanding. 
\OurMethod's strong generalizability with limited training data is also confirmed by its performance on the zero-shot generalization of \GrailQ\nop{, \ie, \OurMethod with BERT-base improves the previous best zero-shot F1 by 2.2\%}. 
Our method also shows great flexibility in accommodating different LMs and a reliable return from model size---using increasingly larger LMs yields monotonically improved results across the board, with T5-3B setting the new state of the art on all datasets.
One interesting observation is that \OurMethod slightly underperforms on the \iid\ subset of \GrailQ. 
It turns out that, because the discriminative task is much easier for LMs to learn than the generative task, \OurMethod converges very fast (at most two epochs) and gets fewer training steps for overfitting the \iid\ setting, in exchange for better non-\iid\ generalization. 
The strong performance on \WebQSP, an \iid\ dataset, further supports this observation.
\end{remark}

\begin{remark}[In-context learning results.]
For the first time, we show the feasibility of effective few-shot KBQA with LLMs. 
On \GrailQ, \OurMethod with Codex achieves an overall F1 of 56.3\% only with 10 training examples. 
Though there is still a gap to the fine-tuning results, it is still impressive, especially considering the massive meaning space of the KB. 
On \GraphQ, \OurMethod with Codex even outperforms ArcaneQA with 10 training examples. 
This further confirms that \OurMethod is particularly strong in generalizing to new environments with limited training data. 
On \WebQSP, \OurMethod trails behind fine-tuning methods when only using 10 training examples, however, increasing the size of the pool for retrieval can significantly boost the performance, which is expected given \WebQSP's \iid\ nature. While for non-\iid datasets like \GrailQ and \GraphQ, the gain from more training examples is marginal.
\end{remark}

Fine-grained performance decomposition by question complexity can be found in~\autoref{appendix:decomposition}, which show that \OurMethod works well across questions of different complexity.

\begin{table*}[!th]
\centering
\small
\resizebox{0.8\textwidth}{!}{
\begin{tabular}{p{0.1\textwidth}p{0.9\textwidth}}
\toprule
\textbf{Question \uppercase\expandafter{\romannumeral1}} & \textit{``neil leslie diamond composed what tv song?"} \tabularnewline
\textbf{Pangu} & \fl{(AND tv.tv\_song (JOIN music.composition.composer m.015\_30))} (\textcolor{teal}{\Checkmark})\\
\textbf{ArcaneQA} & \fl{(AND music.recording (\textcolor{red}{JOIN} music.recording.song (JOIN music.composition.composer m.015\_30)))} (\textcolor{red}{\XSolidBrush})\\
\textbf{ArcaneQA}$^\triangle$ & \fl{(JOIN music.composition.composer m.015\_30) (\textcolor{red}{JOIN} music.recording.song \#0) (AND music.recording \#1)} \\
\midrule
\textbf{Question \uppercase\expandafter{\romannumeral2}} & \textit{``which software falls into both continuous integration and build automation genres?"}\tabularnewline
\textbf{Pangu} & \fl{(AND computer.software (AND (JOIN computer.software.software\_genre m.05vvqy) (JOIN computer.software.software\_genre m.0h2vrf)))} (\textcolor{teal}{\Checkmark})\\
\textbf{ArcaneQA} & \fl{(\textcolor{red}{AND} computer.software (JOIN computer.software.software\_genre m.05vvqy))} (\textcolor{red}{\XSolidBrush})\\
\textbf{ArcaneQA}$^\triangle$ & \fl{(JOIN computer.software.software\_genre m.05vvqy) (\textcolor{red}{AND} computer.software \#0)} \\
\bottomrule
\end{tabular}
}
\vspace{-1em}
\caption{Two representative examples that \OurMethod succeeds while ArcaneQA fails, both w/ BERT-base. $^\triangle$ denotes the original order of the decoder's output. The first incorrect token predicted by ArcaneQA is marked in \textcolor{red}{red}.}
\vspace{-0.5em}
\label{table:case}
\vspace{-5pt}
\end{table*}

\subsection{Sample Efficiency Analysis}
\label{sec:sample_efficiency}

\begin{figure}[t]
    \centering
    \includegraphics[width=\linewidth]{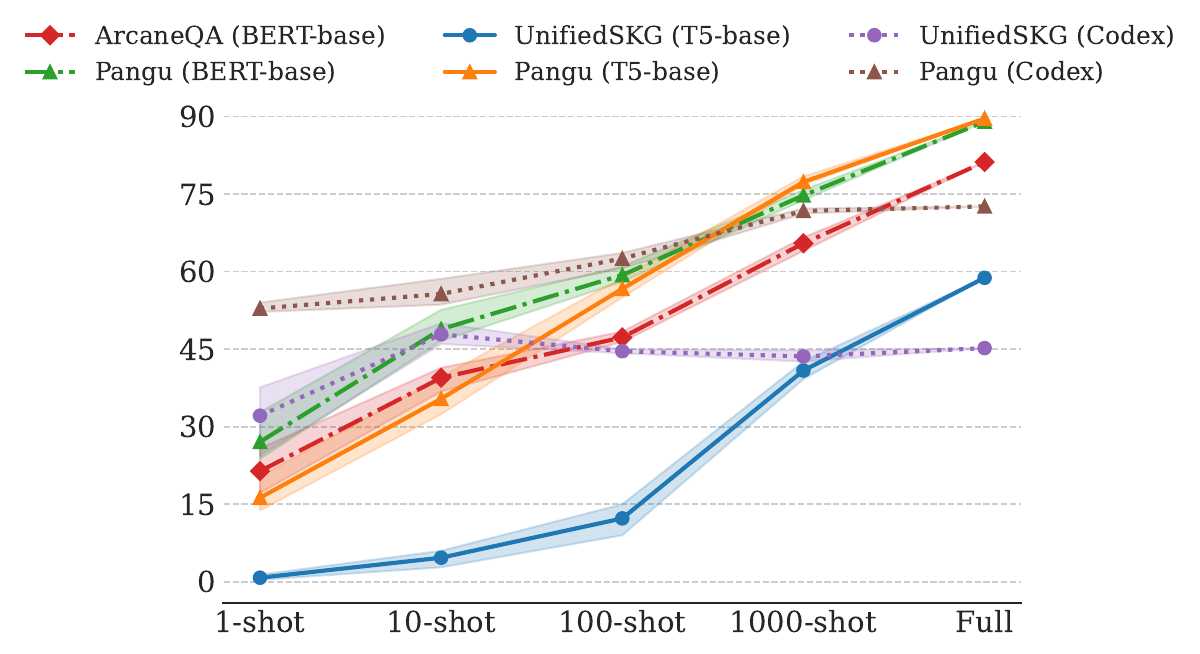}
    \caption{Sample efficiency results. We conduct three runs with different training examples and show the mean EM; shaded areas denote max/min.}
    \vspace{.5em}
    \label{fig:sample}
\end{figure}

Intuitively, by using LMs for discrimination instead of generation, the task becomes easier for LMs and thus improves their sample efficiency.
Our sample efficiency experiments in~\autoref{fig:sample} confirm this hypothesis.
We downsample \GrailQ's training data and randomly sample 1, 10, 100, and 1,000 training examples and report the results on 500 random dev examples.
We compare \OurMethod with ArcaneQA and UnifiedSKG using the same LMs. 
We use oracle entity linking to have a more direct comparison with UnifiedSKG (though UnifiedSKG still has an unfair advantage as previously mentioned). 
In addition, we also include \OurMethod with Codex and use the downsampled training set as the pool for retrieval. 
First, we observe that, when both using T5-base, UnifiedSKG significantly underperforms \OurMethod. 
The main reason is that most predicted plans by UnifiedSKG are invalid in the low-data regime. 
ArcaneQA uses constrained decoding to alleviate this issue, but still consistently underperforms \OurMethod when both using BERT-base. 
For in-context learning using Codex, \OurMethod achieves an EM of over 50\% with only one training instance. 
It consistently outperforms all fine-tuning models under low-data settings (\ie, less than 1,000 training examples). 
Compared with UnifiedSKG, \OurMethod shows both stronger performance and better robustness against different training data selections.

\subsection{\OurMethod vs.\ Constrained Decoding}
\label{sec:constrained_decoding}

\begin{figure}[t]
    \centering
    \includegraphics[width=\linewidth]{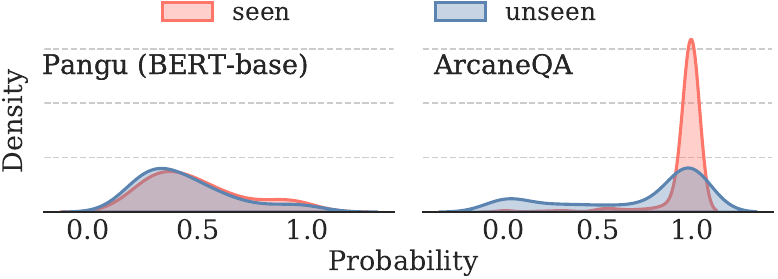}
    \caption{Distribution of the probabilities assigned to predicted programs that are seen and unseen during training. We use kernal density smoothing for better visualization, so the $x$-axis goes over 1.0.}
    \label{fig:confidence}
    \vspace{.5em}
\end{figure}

To better understand \OurMethod's advantage over generation-based methods, we compare \OurMethod with ArcaneQA.
ArcaneQA is the only open-source baseline that uses constrained decoding to enforce the validity of predicted plans.
There are two main reasons for \OurMethod's superiority. First, though constrained decoding can also help ensure plan validity, the autoregressive decoder operates with token-level local normalization and thus lacks a global view. As a result, local failures may break its predictions. For example, a wrong local prediction (e.g., function name) by ArcaneQA leads to catastrophic errors (\autoref{table:case}). 
By evaluating candidate \textit{plans} instead of candidate \textit{tokens}, \OurMethod has a more global view and is less likely to make such local errors.
Second, \OurMethod is less susceptible to overfitting and thus achieves better performance in non-\iid\ settings. \OurMethod does not learn to generate a plan; instead, it learns to evaluate the plausibility of utterance-plan pairs.
Such knowledge is more transferable.
An interesting observation is shown in~\autoref{fig:confidence}, where \OurMethod's output probability distributions are consistent across programs seen and unseen in training. \nop{\footnote{\OurMethod appears to be under-confident. The reason is that it is only trained for ranking with logits.}}
While for ArcaneQA, there is a drastic shift from seen to unseen. This is also consistent with the finding that autoregressive models tend to overfit seen structures during training by \citet{bogin2022unobserved}.
It makes non-\iid\ generalization more difficult.

We also conduct an error analysis in~\autoref{appendix:error}, which sheds some light on future improvements.

\section{Conclusions}

In this paper, we proposed to capitalize on the discriminative ability of language models (LMs) instead of their generative ability for grounded language understanding.
Building on this proposal, we proposed a generic framework, \OurMethod, which consists of a symbolic agent and a neural LM working in a concerted fashion and creates a better separation between the realm of the neural and the symbolic.
This work opens the door for developing versatile and sample-efficient grounded language understanding systems that fully capitalize on the language understanding ability of LMs while avoiding their limitations.
It also sheds light on developing better neuro-symbolic systems in general.
\section*{Limitations}
Despite the strong performance of \OurMethod, we identify several limitations that call for further improvement. 
The first major limitation lies in efficiency. Because \OurMethod requires an LM to iteratively score candidate plans, it is resource-consuming in terms of both time and computing. Compared with ArcaneQA, which efficiently handles complex questions in KBQA, \OurMethod is about twice slower for both training and inference and consumes about twice as much GPU memory when using the same LM. Concretely, to predict a plan of $L$ tokens, generation-based methods involve using an LM to do $L$ forward passes. For \OurMethod, the number of forward passes is proportional to the number of candidate plans, which can range widely. In the future, algorithms with complexity better than $O(N)$, $N$ being the number of candidate plans, are desired to find the top-$K$ candidates.

Second, though \OurMethod has shown some promising results with Codex, the true potential of enabling few-shot grounded language understanding with \OurMethod has yet to be realized. We only experiment with a straightforward scoring function and have not experimented with different prompt designs systematically. In the future, we plan to try different prompt designs, retrievers, and scoring functions, including using latest techniques like chain of thought~\cite{wei2022chain}.

Third, though orthogonal to the general framework of our proposal, in our current instantiation, we assume gold plans for training. 
However, gold plans can be expensive to collect for some environments. 
Exploring fine-tuning LMs with weak supervision can be an interesting direction.
In addition to proposing candidate plans to the LM, the agent may also respond to the LM with rewards based on its decisions~\cite{liang-etal-2017-neural}.

Finally, in this paper, one important merit of \OurMethod, controllability, is under-explored, because it is not very necessary for KBQA. While for tasks like text-to-SQL parsing, controllability is a highly desirable property. Intruders may manipulate text-to-SQL models to launch database attacks via SQL injection~\cite{peng2022security}. With \OurMethod, we can easily get rid of malicious SQL operations in candidate enumeration. However, for generation-based methods, such controls are hard to achieve during generation because the decoding process can be shortsighted; it is difficult to tell whether the current prediction will lead to a malicious operation several steps later. In the future, we will explore \OurMethod's controllability on more different tasks.
\section*{Acknowledgements}
The authors would like to thank Percy Liang, Jiawei Han, Jonathan Berant, Huan Sun, and other colleagues from the
OSU NLP group for their valuable feedback. The authors would also like to thank Yiheng Shu for sharing their entity linking results and Tianbao Xie for clarifications on UnifiedSKG. This research was supported in part by ARL W911NF2220144, NSF OAC 2112606, and Ohio Supercomputer Center~\cite{OhioSupercomputerCenter1987}.

\bibliography{anthology,custom}

\begin{thebibliography}{66}
\expandafter\ifx\csname natexlab\endcsname\relax\def\natexlab#1{#1}\fi

\bibitem[{Ahn et~al.(2022)Ahn, Brohan, Brown, Chebotar, Cortes, David, Finn,
  Gopalakrishnan, Hausman, Herzog et~al.}]{ahn2022can}
Michael Ahn, Anthony Brohan, Noah Brown, Yevgen Chebotar, Omar Cortes, Byron
  David, Chelsea Finn, Keerthana Gopalakrishnan, Karol Hausman, Alex Herzog,
  et~al. 2022.
\newblock Do as i can, not as i say: Grounding language in robotic affordances.
\newblock \emph{arXiv preprint arXiv:2204.01691}.

\bibitem[{Andreas et~al.(2020)Andreas, Bufe, Burkett, Chen, Clausman, Crawford,
  Crim, DeLoach, Dorner, Eisner, Fang, Guo, Hall, Hayes, Hill, Ho, Iwaszuk,
  Jha, Klein, Krishnamurthy, Lanman, Liang, Lin, Lintsbakh, McGovern,
  Nisnevich, Pauls, Petters, Read, Roth, Roy, Rusak, Short, Slomin, Snyder,
  Striplin, Su, Tellman, Thomson, Vorobev, Witoszko, Wolfe, Wray, Zhang, and
  Zotov}]{andreas-etal-2020-task}
Jacob Andreas, John Bufe, David Burkett, Charles Chen, Josh Clausman, Jean
  Crawford, Kate Crim, Jordan DeLoach, Leah Dorner, Jason Eisner, Hao Fang,
  Alan Guo, David Hall, Kristin Hayes, Kellie Hill, Diana Ho, Wendy Iwaszuk,
  Smriti Jha, Dan Klein, Jayant Krishnamurthy, Theo Lanman, Percy Liang,
  Christopher~H. Lin, Ilya Lintsbakh, Andy McGovern, Aleksandr Nisnevich, Adam
  Pauls, Dmitrij Petters, Brent Read, Dan Roth, Subhro Roy, Jesse Rusak, Beth
  Short, Div Slomin, Ben Snyder, Stephon Striplin, Yu~Su, Zachary Tellman, Sam
  Thomson, Andrei Vorobev, Izabela Witoszko, Jason Wolfe, Abby Wray, Yuchen
  Zhang, and Alexander Zotov. 2020.
\newblock \href {https://doi.org/10.1162/tacl_a_00333} {Task-oriented dialogue
  as dataflow synthesis}.
\newblock \emph{Transactions of the Association for Computational Linguistics},
  8:556--571.

\bibitem[{Austin et~al.(2021)Austin, Odena, Nye, Bosma, Michalewski, Dohan,
  Jiang, Cai, Terry, Le, and Sutton}]{Austin2021ProgramSW}
Jacob Austin, Augustus Odena, Maxwell~I. Nye, Maarten Bosma, Henryk
  Michalewski, David Dohan, Ellen Jiang, Carrie~J. Cai, Michael Terry, Quoc~V.
  Le, and Charles Sutton. 2021.
\newblock \href {http://arxiv.org/abs/2108.07732} {Program synthesis with large
  language models}.
\newblock \emph{CoRR}, abs/2108.07732.

\bibitem[{Bahdanau et~al.(2015)Bahdanau, Cho, and Bengio}]{bahdanau2014neural}
Dzmitry Bahdanau, Kyunghyun Cho, and Yoshua Bengio. 2015.
\newblock \href {http://arxiv.org/abs/1409.0473} {Neural machine translation by
  jointly learning to align and translate}.
\newblock In \emph{3rd International Conference on Learning Representations,
  {ICLR} 2015, San Diego, CA, USA, May 7-9, 2015, Conference Track
  Proceedings}.

\bibitem[{Berant et~al.(2013)Berant, Chou, Frostig, and
  Liang}]{berant-etal-2013-semantic}
Jonathan Berant, Andrew Chou, Roy Frostig, and Percy Liang. 2013.
\newblock \href {https://aclanthology.org/D13-1160} {Semantic parsing on
  {F}reebase from question-answer pairs}.
\newblock In \emph{Proceedings of the 2013 Conference on Empirical Methods in
  Natural Language Processing}, pages 1533--1544, Seattle, Washington, USA.
  Association for Computational Linguistics.

\bibitem[{Bogin et~al.(2022)Bogin, Gupta, and Berant}]{bogin2022unobserved}
Ben Bogin, Shivanshu Gupta, and Jonathan Berant. 2022.
\newblock Unobserved local structures make compositional generalization hard.
\newblock \emph{arXiv preprint arXiv:2201.05899}.

\bibitem[{Bollacker et~al.(2008)Bollacker, Evans, Paritosh, Sturge, and
  Taylor}]{bollacker2008freebase}
Kurt~D. Bollacker, Colin Evans, Praveen~K. Paritosh, Tim Sturge, and Jamie
  Taylor. 2008.
\newblock \href {https://doi.org/10.1145/1376616.1376746} {Freebase: a
  collaboratively created graph database for structuring human knowledge}.
\newblock In \emph{Proceedings of the {ACM} {SIGMOD} International Conference
  on Management of Data, {SIGMOD} 2008, Vancouver, BC, Canada, June 10-12,
  2008}, pages 1247--1250. {ACM}.

\bibitem[{Brown et~al.(2020)Brown, Mann, Ryder, Subbiah, Kaplan, Dhariwal,
  Neelakantan, Shyam, Sastry, Askell, Agarwal, Herbert{-}Voss, Krueger,
  Henighan, Child, Ramesh, Ziegler, Wu, Winter, Hesse, Chen, Sigler, Litwin,
  Gray, Chess, Clark, Berner, McCandlish, Radford, Sutskever, and
  Amodei}]{brown2020language}
Tom~B. Brown, Benjamin Mann, Nick Ryder, Melanie Subbiah, Jared Kaplan,
  Prafulla Dhariwal, Arvind Neelakantan, Pranav Shyam, Girish Sastry, Amanda
  Askell, Sandhini Agarwal, Ariel Herbert{-}Voss, Gretchen Krueger, Tom
  Henighan, Rewon Child, Aditya Ramesh, Daniel~M. Ziegler, Jeffrey Wu, Clemens
  Winter, Christopher Hesse, Mark Chen, Eric Sigler, Mateusz Litwin, Scott
  Gray, Benjamin Chess, Jack Clark, Christopher Berner, Sam McCandlish, Alec
  Radford, Ilya Sutskever, and Dario Amodei. 2020.
\newblock \href
  {https://proceedings.neurips.cc/paper/2020/hash/1457c0d6bfcb4967418bfb8ac142f64a-Abstract.html}
  {Language models are few-shot learners}.
\newblock In \emph{Advances in Neural Information Processing Systems 33: Annual
  Conference on Neural Information Processing Systems 2020, NeurIPS 2020,
  December 6-12, 2020, virtual}.

\bibitem[{Cao et~al.(2022)Cao, Shi, Yao, Lv, Yu, Hou, Li, Liu, and
  Xiao}]{cao-etal-2022-program}
Shulin Cao, Jiaxin Shi, Zijun Yao, Xin Lv, Jifan Yu, Lei Hou, Juanzi Li,
  Zhiyuan Liu, and Jinghui Xiao. 2022.
\newblock \href {https://doi.org/10.18653/v1/2022.acl-long.559} {Program
  transfer for answering complex questions over knowledge bases}.
\newblock In \emph{Proceedings of the 60th Annual Meeting of the Association
  for Computational Linguistics (Volume 1: Long Papers)}, pages 8128--8140,
  Dublin, Ireland. Association for Computational Linguistics.

\bibitem[{Cao et~al.(2007)Cao, Qin, Liu, Tsai, and Li}]{cao2007learning}
Zhe Cao, Tao Qin, Tie{-}Yan Liu, Ming{-}Feng Tsai, and Hang Li. 2007.
\newblock \href {https://doi.org/10.1145/1273496.1273513} {Learning to rank:
  from pairwise approach to listwise approach}.
\newblock In \emph{Machine Learning, Proceedings of the Twenty-Fourth
  International Conference {(ICML} 2007), Corvallis, Oregon, USA, June 20-24,
  2007}, volume 227 of \emph{{ACM} International Conference Proceeding Series},
  pages 129--136. {ACM}.

\bibitem[{Center(1987)}]{OhioSupercomputerCenter1987}
Ohio~Supercomputer Center. 1987.
\newblock \href {http://osc.edu/ark:/19495/f5s1ph73} {{Ohio Supercomputer
  Center}}.

\bibitem[{Chandu et~al.(2021)Chandu, Bisk, and
  Black}]{chandu-etal-2021-grounding}
Khyathi~Raghavi Chandu, Yonatan Bisk, and Alan~W Black. 2021.
\newblock \href {https://doi.org/10.18653/v1/2021.findings-acl.375} {Grounding
  {`}grounding{'} in {NLP}}.
\newblock In \emph{Findings of the Association for Computational Linguistics:
  ACL-IJCNLP 2021}, pages 4283--4305, Online. Association for Computational
  Linguistics.

\bibitem[{Chen et~al.(2021{\natexlab{a}})Chen, Tworek, Jun, Yuan,
  de~Oliveira~Pinto, Kaplan, Edwards, Burda, Joseph, Brockman, Ray, Puri,
  Krueger, Petrov, Khlaaf, Sastry, Mishkin, Chan, Gray, Ryder, Pavlov, Power,
  Kaiser, Bavarian, Winter, Tillet, Such, Cummings, Plappert, Chantzis, Barnes,
  Herbert{-}Voss, Guss, Nichol, Paino, Tezak, Tang, Babuschkin, Balaji, Jain,
  Saunders, Hesse, Carr, Leike, Achiam, Misra, Morikawa, Radford, Knight,
  Brundage, Murati, Mayer, Welinder, McGrew, Amodei, McCandlish, Sutskever, and
  Zaremba}]{chen2021codex}
Mark Chen, Jerry Tworek, Heewoo Jun, Qiming Yuan, Henrique~Ponde
  de~Oliveira~Pinto, Jared Kaplan, Harrison Edwards, Yuri Burda, Nicholas
  Joseph, Greg Brockman, Alex Ray, Raul Puri, Gretchen Krueger, Michael Petrov,
  Heidy Khlaaf, Girish Sastry, Pamela Mishkin, Brooke Chan, Scott Gray, Nick
  Ryder, Mikhail Pavlov, Alethea Power, Lukasz Kaiser, Mohammad Bavarian,
  Clemens Winter, Philippe Tillet, Felipe~Petroski Such, Dave Cummings,
  Matthias Plappert, Fotios Chantzis, Elizabeth Barnes, Ariel Herbert{-}Voss,
  William~Hebgen Guss, Alex Nichol, Alex Paino, Nikolas Tezak, Jie Tang, Igor
  Babuschkin, Suchir Balaji, Shantanu Jain, William Saunders, Christopher
  Hesse, Andrew~N. Carr, Jan Leike, Joshua Achiam, Vedant Misra, Evan Morikawa,
  Alec Radford, Matthew Knight, Miles Brundage, Mira Murati, Katie Mayer, Peter
  Welinder, Bob McGrew, Dario Amodei, Sam McCandlish, Ilya Sutskever, and
  Wojciech Zaremba. 2021{\natexlab{a}}.
\newblock \href {http://arxiv.org/abs/2107.03374} {Evaluating large language
  models trained on code}.
\newblock \emph{CoRR}, abs/2107.03374.

\bibitem[{Chen et~al.(2021{\natexlab{b}})Chen, Liu, Yu, Lin, Lou, and
  Jiang}]{chen-etal-2021-retrack}
Shuang Chen, Qian Liu, Zhiwei Yu, Chin-Yew Lin, Jian-Guang Lou, and Feng Jiang.
  2021{\natexlab{b}}.
\newblock \href {https://doi.org/10.18653/v1/2021.acl-demo.39} {{R}e{T}ra{C}k:
  A flexible and efficient framework for knowledge base question answering}.
\newblock In \emph{Proceedings of the 59th Annual Meeting of the Association
  for Computational Linguistics and the 11th International Joint Conference on
  Natural Language Processing: System Demonstrations}, pages 325--336, Online.
  Association for Computational Linguistics.

\bibitem[{Cheng et~al.(2022)Cheng, Xie, Shi, Li, Nadkarni, Hu, Xiong, Radev,
  Ostendorf, Zettlemoyer, Smith, and Yu}]{cheng2022binding}
Zhoujun Cheng, Tianbao Xie, Peng Shi, Chengzu Li, Rahul Nadkarni, Yushi Hu,
  Caiming Xiong, Dragomir Radev, Mari Ostendorf, Luke Zettlemoyer, Noah~A.
  Smith, and Tao Yu. 2022.
\newblock \href {https://doi.org/10.48550/arXiv.2210.02875} {Binding language
  models in symbolic languages}.
\newblock \emph{CoRR}, abs/2210.02875.

\bibitem[{Cho et~al.(2014)Cho, van Merri{\"e}nboer, Gulcehre, Bahdanau,
  Bougares, Schwenk, and Bengio}]{cho-etal-2014-learning}
Kyunghyun Cho, Bart van Merri{\"e}nboer, Caglar Gulcehre, Dzmitry Bahdanau,
  Fethi Bougares, Holger Schwenk, and Yoshua Bengio. 2014.
\newblock \href {https://doi.org/10.3115/v1/D14-1179} {Learning phrase
  representations using {RNN} encoder{--}decoder for statistical machine
  translation}.
\newblock In \emph{Proceedings of the 2014 Conference on Empirical Methods in
  Natural Language Processing ({EMNLP})}, pages 1724--1734, Doha, Qatar.
  Association for Computational Linguistics.

\bibitem[{Codd(1970)}]{codd1970relational}
E.~F. Codd. 1970.
\newblock \href {https://doi.org/10.1145/362384.362685} {A relational model of
  data for large shared data banks}.
\newblock \emph{Commun. {ACM}}, 13(6):377--387.

\bibitem[{Das et~al.(2021)Das, Zaheer, Thai, Godbole, Perez, Lee, Tan,
  Polymenakos, and McCallum}]{das-etal-2021-case}
Rajarshi Das, Manzil Zaheer, Dung Thai, Ameya Godbole, Ethan Perez, Jay~Yoon
  Lee, Lizhen Tan, Lazaros Polymenakos, and Andrew McCallum. 2021.
\newblock \href {https://doi.org/10.18653/v1/2021.emnlp-main.755} {Case-based
  reasoning for natural language queries over knowledge bases}.
\newblock In \emph{Proceedings of the 2021 Conference on Empirical Methods in
  Natural Language Processing}, pages 9594--9611, Online and Punta Cana,
  Dominican Republic. Association for Computational Linguistics.

\bibitem[{Deng et~al.(2021)Deng, Awadallah, Meek, Polozov, Sun, and
  Richardson}]{deng-etal-2021-structure}
Xiang Deng, Ahmed~Hassan Awadallah, Christopher Meek, Oleksandr Polozov, Huan
  Sun, and Matthew Richardson. 2021.
\newblock \href {https://doi.org/10.18653/v1/2021.naacl-main.105}
  {Structure-grounded pretraining for text-to-{SQL}}.
\newblock In \emph{Proceedings of the 2021 Conference of the North American
  Chapter of the Association for Computational Linguistics: Human Language
  Technologies}, pages 1337--1350, Online. Association for Computational
  Linguistics.

\bibitem[{Devlin et~al.(2019)Devlin, Chang, Lee, and
  Toutanova}]{devlin-etal-2019-bert}
Jacob Devlin, Ming-Wei Chang, Kenton Lee, and Kristina Toutanova. 2019.
\newblock \href {https://doi.org/10.18653/v1/N19-1423} {{BERT}: Pre-training of
  deep bidirectional transformers for language understanding}.
\newblock In \emph{Proceedings of the 2019 Conference of the North {A}merican
  Chapter of the Association for Computational Linguistics: Human Language
  Technologies, Volume 1 (Long and Short Papers)}, pages 4171--4186,
  Minneapolis, Minnesota. Association for Computational Linguistics.

\bibitem[{Dong et~al.(2017)Dong, Mallinson, Reddy, and
  Lapata}]{dong-etal-2017-learning}
Li~Dong, Jonathan Mallinson, Siva Reddy, and Mirella Lapata. 2017.
\newblock \href {https://doi.org/10.18653/v1/D17-1091} {Learning to paraphrase
  for question answering}.
\newblock In \emph{Proceedings of the 2017 Conference on Empirical Methods in
  Natural Language Processing}, pages 875--886, Copenhagen, Denmark.
  Association for Computational Linguistics.

\bibitem[{Gu et~al.(2021)Gu, Kase, Vanni, Sadler, Liang, Yan, and
  Su}]{gu2021beyond}
Yu~Gu, Sue Kase, Michelle Vanni, Brian~M. Sadler, Percy Liang, Xifeng Yan, and
  Yu~Su. 2021.
\newblock \href {https://doi.org/10.1145/3442381.3449992} {Beyond {I.I.D.:}
  three levels of generalization for question answering on knowledge bases}.
\newblock In \emph{{WWW} '21: The Web Conference 2021, Virtual Event /
  Ljubljana, Slovenia, April 19-23, 2021}, pages 3477--3488. {ACM} / {IW3C2}.

\bibitem[{Gu et~al.(2022)Gu, Pahuja, Cheng, and Su}]{gu2022knowledge}
Yu~Gu, Vardaan Pahuja, Gong Cheng, and Yu~Su. 2022.
\newblock Knowledge base question answering: A semantic parsing perspective.
\newblock In \emph{4th Conference on Automated Knowledge Base Construction}.

\bibitem[{Gu and Su(2022)}]{gu-su-2022-arcaneqa}
Yu~Gu and Yu~Su. 2022.
\newblock \href {https://aclanthology.org/2022.coling-1.148} {{A}rcane{QA}:
  Dynamic program induction and contextualized encoding for knowledge base
  question answering}.
\newblock In \emph{Proceedings of the 29th International Conference on
  Computational Linguistics}, pages 1718--1731, Gyeongju, Republic of Korea.
  International Committee on Computational Linguistics.

\bibitem[{Gur et~al.(2022)Gur, Nachum, Miao, Safdari, Huang, Chowdhery, Narang,
  Fiedel, and Faust}]{Gur2022UnderstandingHW}
Izzeddin Gur, Ofir Nachum, Yingjie Miao, Mustafa Safdari, Austin Huang,
  Aakanksha Chowdhery, Sharan Narang, Noah Fiedel, and Aleksandra Faust. 2022.
\newblock \href {https://doi.org/10.48550/arXiv.2210.03945} {Understanding
  {HTML} with large language models}.
\newblock \emph{CoRR}, abs/2210.03945.

\bibitem[{Hwang et~al.(2019)Hwang, Yim, Park, and Seo}]{hwang2019comprehensive}
Wonseok Hwang, Jinyeung Yim, Seunghyun Park, and Minjoon Seo. 2019.
\newblock \href {http://arxiv.org/abs/1902.01069} {A comprehensive exploration
  on wikisql with table-aware word contextualization}.
\newblock \emph{CoRR}, abs/1902.01069.

\bibitem[{Izacard and Grave(2021)}]{izacard-grave-2021-leveraging}
Gautier Izacard and Edouard Grave. 2021.
\newblock \href {https://doi.org/10.18653/v1/2021.eacl-main.74} {Leveraging
  passage retrieval with generative models for open domain question answering}.
\newblock In \emph{Proceedings of the 16th Conference of the European Chapter
  of the Association for Computational Linguistics: Main Volume}, pages
  874--880, Online. Association for Computational Linguistics.

\bibitem[{Jain et~al.(2022)Jain, Vaidyanath, Iyer, Natarajan, Parthasarathy,
  Rajamani, and Sharma}]{Jain2021JigsawLL}
Naman Jain, Skanda Vaidyanath, Arun~Shankar Iyer, Nagarajan Natarajan, Suresh
  Parthasarathy, Sriram~K. Rajamani, and Rahul Sharma. 2022.
\newblock \href {https://doi.org/10.1145/3510003.3510203} {Jigsaw: Large
  language models meet program synthesis}.
\newblock In \emph{44th {IEEE/ACM} 44th International Conference on Software
  Engineering, {ICSE} 2022, Pittsburgh, PA, USA, May 25-27, 2022}, pages
  1219--1231. {ACM}.

\bibitem[{Karpukhin et~al.(2020)Karpukhin, Oguz, Min, Lewis, Wu, Edunov, Chen,
  and Yih}]{karpukhin-etal-2020-dense}
Vladimir Karpukhin, Barlas Oguz, Sewon Min, Patrick Lewis, Ledell Wu, Sergey
  Edunov, Danqi Chen, and Wen-tau Yih. 2020.
\newblock \href {https://doi.org/10.18653/v1/2020.emnlp-main.550} {Dense
  passage retrieval for open-domain question answering}.
\newblock In \emph{Proceedings of the 2020 Conference on Empirical Methods in
  Natural Language Processing (EMNLP)}, pages 6769--6781, Online. Association
  for Computational Linguistics.

\bibitem[{Lan and Jiang(2020)}]{lan-jiang-2020-query}
Yunshi Lan and Jing Jiang. 2020.
\newblock \href {https://doi.org/10.18653/v1/2020.acl-main.91} {Query graph
  generation for answering multi-hop complex questions from knowledge bases}.
\newblock In \emph{Proceedings of the 58th Annual Meeting of the Association
  for Computational Linguistics}, pages 969--974, Online. Association for
  Computational Linguistics.

\bibitem[{Li et~al.(2020)Li, Min, Iyer, Mehdad, and
  Yih}]{li-etal-2020-efficient}
Belinda~Z. Li, Sewon Min, Srinivasan Iyer, Yashar Mehdad, and Wen-tau Yih.
  2020.
\newblock \href {https://doi.org/10.18653/v1/2020.emnlp-main.522} {Efficient
  one-pass end-to-end entity linking for questions}.
\newblock In \emph{Proceedings of the 2020 Conference on Empirical Methods in
  Natural Language Processing (EMNLP)}, pages 6433--6441, Online. Association
  for Computational Linguistics.

\bibitem[{Li et~al.(2019)Li, Yatskar, Yin, Hsieh, and Chang}]{visualbert}
Liunian~Harold Li, Mark Yatskar, Da~Yin, Cho-Jui Hsieh, and Kai-Wei Chang.
  2019.
\newblock \href {https://doi.org/10.48550/ARXIV.1908.03557} {Visualbert: A
  simple and performant baseline for vision and language}.

\bibitem[{Li et~al.(2022)Li, Choi, Chung, Kushman, Schrittwieser, Leblond,
  Eccles, Keeling, Gimeno, Lago, Hubert, Choy, de~Masson~d’Autume,
  Babuschkin, Chen, Huang, Welbl, Gowal, Cherepanov, Molloy, Mankowitz, Robson,
  Kohli, de~Freitas, Kavukcuoglu, and Vinyals}]{Li2022CompetitionLevelCG}
Yujia Li, David Choi, Junyoung Chung, Nate Kushman, Julian Schrittwieser, Rémi
  Leblond, Tom Eccles, James Keeling, Felix Gimeno, Agustin~Dal Lago, Thomas
  Hubert, Peter Choy, Cyprien de~Masson~d’Autume, Igor Babuschkin, Xinyun
  Chen, Po-Sen Huang, Johannes Welbl, Sven Gowal, Alexey Cherepanov, James
  Molloy, Daniel~J. Mankowitz, Esme~Sutherland Robson, Pushmeet Kohli, Nando
  de~Freitas, Koray Kavukcuoglu, and Oriol Vinyals. 2022.
\newblock \href {https://doi.org/10.1126/science.abq1158} {Competition-level
  code generation with alphacode}.
\newblock \emph{Science}, 378(6624):1092--1097.

\bibitem[{Liang et~al.(2017)Liang, Berant, Le, Forbus, and
  Lao}]{liang-etal-2017-neural}
Chen Liang, Jonathan Berant, Quoc Le, Kenneth~D. Forbus, and Ni~Lao. 2017.
\newblock \href {https://doi.org/10.18653/v1/P17-1003} {Neural symbolic
  machines: Learning semantic parsers on {F}reebase with weak supervision}.
\newblock In \emph{Proceedings of the 55th Annual Meeting of the Association
  for Computational Linguistics (Volume 1: Long Papers)}, pages 23--33,
  Vancouver, Canada. Association for Computational Linguistics.

\bibitem[{Liang et~al.(2022)Liang, Bommasani, Lee, Tsipras, Soylu, Yasunaga,
  Zhang, Narayanan, Wu, Kumar, Newman, Yuan, Yan, Zhang, Cosgrove, Manning,
  R{\'{e}}, Acosta{-}Navas, Hudson, Zelikman, Durmus, Ladhak, Rong, Ren, Yao,
  Wang, Santhanam, Orr, Zheng, Y{\"{u}}ksekg{\"{o}}n{\"{u}}l, Suzgun, Kim,
  Guha, Chatterji, Khattab, Henderson, Huang, Chi, Xie, Santurkar, Ganguli,
  Hashimoto, Icard, Zhang, Chaudhary, Wang, Li, Mai, Zhang, and
  Koreeda}]{liang2022holistic}
Percy Liang, Rishi Bommasani, Tony Lee, Dimitris Tsipras, Dilara Soylu,
  Michihiro Yasunaga, Yian Zhang, Deepak Narayanan, Yuhuai Wu, Ananya Kumar,
  Benjamin Newman, Binhang Yuan, Bobby Yan, Ce~Zhang, Christian Cosgrove,
  Christopher~D. Manning, Christopher R{\'{e}}, Diana Acosta{-}Navas, Drew~A.
  Hudson, Eric Zelikman, Esin Durmus, Faisal Ladhak, Frieda Rong, Hongyu Ren,
  Huaxiu Yao, Jue Wang, Keshav Santhanam, Laurel~J. Orr, Lucia Zheng, Mert
  Y{\"{u}}ksekg{\"{o}}n{\"{u}}l, Mirac Suzgun, Nathan Kim, Neel Guha,
  Niladri~S. Chatterji, Omar Khattab, Peter Henderson, Qian Huang, Ryan Chi,
  Sang~Michael Xie, Shibani Santurkar, Surya Ganguli, Tatsunori Hashimoto,
  Thomas Icard, Tianyi Zhang, Vishrav Chaudhary, William Wang, Xuechen Li,
  Yifan Mai, Yuhui Zhang, and Yuta Koreeda. 2022.
\newblock \href {https://doi.org/10.48550/arXiv.2211.09110} {Holistic
  evaluation of language models}.
\newblock \emph{CoRR}, abs/2211.09110.

\bibitem[{Liu et~al.(2022)Liu, Yavuz, Meng, Radev, Xiong, and
  Zhou}]{liu2022uni}
Ye~Liu, Semih Yavuz, Rui Meng, Dragomir Radev, Caiming Xiong, and Yingbo Zhou.
  2022.
\newblock \href {https://doi.org/10.48550/arXiv.2211.05165} {Uni-parser:
  Unified semantic parser for question answering on knowledge base and
  database}.
\newblock \emph{CoRR}, abs/2211.05165.

\bibitem[{Lu et~al.(2019)Lu, Batra, Parikh, and Lee}]{vilbert}
Jiasen Lu, Dhruv Batra, Devi Parikh, and Stefan Lee. 2019.
\newblock \href
  {https://proceedings.neurips.cc/paper/2019/hash/c74d97b01eae257e44aa9d5bade97baf-Abstract.html}
  {Vilbert: Pretraining task-agnostic visiolinguistic representations for
  vision-and-language tasks}.
\newblock In \emph{Advances in Neural Information Processing Systems 32: Annual
  Conference on Neural Information Processing Systems 2019, NeurIPS 2019,
  December 8-14, 2019, Vancouver, BC, Canada}, pages 13--23.

\bibitem[{Minaee et~al.(2021)Minaee, Kalchbrenner, Cambria, Nikzad, Chenaghlu,
  and Gao}]{Minaee2021DeepLB}
Shervin Minaee, Nal Kalchbrenner, Erik Cambria, Narjes Nikzad, Meysam
  Chenaghlu, and Jianfeng Gao. 2021.
\newblock \href {https://doi.org/10.1145/3439726} {Deep learning--based text
  classification: A comprehensive review}.
\newblock \emph{ACM Comput. Surv.}, 54(3).

\bibitem[{Nakano et~al.(2021)Nakano, Hilton, Balaji, Wu, Ouyang, Kim, Hesse,
  Jain, Kosaraju, Saunders, Jiang, Cobbe, Eloundou, Krueger, Button, Knight,
  Chess, and Schulman}]{Nakano2021WebGPTBQ}
Reiichiro Nakano, Jacob Hilton, Suchir Balaji, Jeff Wu, Long Ouyang, Christina
  Kim, Christopher Hesse, Shantanu Jain, Vineet Kosaraju, William Saunders,
  Xu~Jiang, Karl Cobbe, Tyna Eloundou, Gretchen Krueger, Kevin Button, Matthew
  Knight, Benjamin Chess, and John Schulman. 2021.
\newblock \href {http://arxiv.org/abs/2112.09332} {Webgpt: Browser-assisted
  question-answering with human feedback}.
\newblock \emph{CoRR}, abs/2112.09332.

\bibitem[{Peng et~al.(2022)Peng, Zhang, Yang, and Stevenson}]{peng2022security}
Xutan Peng, Yipeng Zhang, Jingfeng Yang, and Mark Stevenson. 2022.
\newblock On the security vulnerabilities of text-to-sql models.
\newblock \emph{arXiv preprint arXiv:2211.15363}.

\bibitem[{Rajkumar et~al.(2022)Rajkumar, Li, and
  Bahdanau}]{Rajkumar2022EvaluatingTT}
Nitarshan Rajkumar, Raymond Li, and Dzmitry Bahdanau. 2022.
\newblock \href {https://doi.org/10.48550/arXiv.2204.00498} {Evaluating the
  text-to-sql capabilities of large language models}.
\newblock \emph{CoRR}, abs/2204.00498.

\bibitem[{Reddy et~al.(2017)Reddy, T{\"a}ckstr{\"o}m, Petrov, Steedman, and
  Lapata}]{reddy-etal-2017-universal}
Siva Reddy, Oscar T{\"a}ckstr{\"o}m, Slav Petrov, Mark Steedman, and Mirella
  Lapata. 2017.
\newblock \href {https://doi.org/10.18653/v1/D17-1009} {Universal semantic
  parsing}.
\newblock In \emph{Proceedings of the 2017 Conference on Empirical Methods in
  Natural Language Processing}, pages 89--101, Copenhagen, Denmark. Association
  for Computational Linguistics.

\bibitem[{Rubin and Berant(2021)}]{rubin-berant-2021-smbop}
Ohad Rubin and Jonathan Berant. 2021.
\newblock \href {https://doi.org/10.18653/v1/2021.naacl-main.29} {{S}m{B}o{P}:
  Semi-autoregressive bottom-up semantic parsing}.
\newblock In \emph{Proceedings of the 2021 Conference of the North American
  Chapter of the Association for Computational Linguistics: Human Language
  Technologies}, pages 311--324, Online. Association for Computational
  Linguistics.

\bibitem[{Scholak et~al.(2021)Scholak, Schucher, and
  Bahdanau}]{scholak-etal-2021-picard}
Torsten Scholak, Nathan Schucher, and Dzmitry Bahdanau. 2021.
\newblock \href {https://doi.org/10.18653/v1/2021.emnlp-main.779} {{PICARD}:
  Parsing incrementally for constrained auto-regressive decoding from language
  models}.
\newblock In \emph{Proceedings of the 2021 Conference on Empirical Methods in
  Natural Language Processing}, pages 9895--9901, Online and Punta Cana,
  Dominican Republic. Association for Computational Linguistics.

\bibitem[{Shah et~al.(2022)Shah, Osi{\'n}ski, brian ichter, and
  Levine}]{shah2022lmnav}
Dhruv Shah, B{\l}a{\.z}ej Osi{\'n}ski, brian ichter, and Sergey Levine. 2022.
\newblock \href {https://openreview.net/forum?id=UW5A3SweAH} {{LM}-nav: Robotic
  navigation with large pre-trained models of language, vision, and action}.
\newblock In \emph{6th Annual Conference on Robot Learning}.

\bibitem[{Shin et~al.(2021)Shin, Lin, Thomson, Chen, Roy, Platanios, Pauls,
  Klein, Eisner, and Van~Durme}]{shin-etal-2021-constrained}
Richard Shin, Christopher Lin, Sam Thomson, Charles Chen, Subhro Roy,
  Emmanouil~Antonios Platanios, Adam Pauls, Dan Klein, Jason Eisner, and
  Benjamin Van~Durme. 2021.
\newblock \href {https://doi.org/10.18653/v1/2021.emnlp-main.608} {Constrained
  language models yield few-shot semantic parsers}.
\newblock In \emph{Proceedings of the 2021 Conference on Empirical Methods in
  Natural Language Processing}, pages 7699--7715, Online and Punta Cana,
  Dominican Republic. Association for Computational Linguistics.

\bibitem[{Shridhar et~al.(2020)Shridhar, Thomason, Gordon, Bisk, Han, Mottaghi,
  Zettlemoyer, and Fox}]{shridhar2020alfred}
Mohit Shridhar, Jesse Thomason, Daniel Gordon, Yonatan Bisk, Winson Han,
  Roozbeh Mottaghi, Luke Zettlemoyer, and Dieter Fox. 2020.
\newblock \href {https://doi.org/10.1109/CVPR42600.2020.01075} {{ALFRED:} {A}
  benchmark for interpreting grounded instructions for everyday tasks}.
\newblock In \emph{2020 {IEEE/CVF} Conference on Computer Vision and Pattern
  Recognition, {CVPR} 2020, Seattle, WA, USA, June 13-19, 2020}, pages
  10737--10746. Computer Vision Foundation / {IEEE}.

\bibitem[{Shu et~al.(2022)Shu, Yu, Li, Karlsson, Ma, Qu, and
  Lin}]{shu2022tiara}
Yiheng Shu, Zhiwei Yu, Yuhan Li, B{\"o}rje~F Karlsson, Tingting Ma, Yuzhong Qu,
  and Chin-Yew Lin. 2022.
\newblock {TIARA}: Multi-grained retrieval for robust question answering over
  large knowledge bases.
\newblock In \emph{Proceedings of the 2022 Conference on Empirical Methods in
  Natural Language Processing}.

\bibitem[{Singh et~al.(2022)Singh, Blukis, Mousavian, Goyal, Xu, Tremblay, Fox,
  Thomason, and Garg}]{singh2022progprompt}
Ishika Singh, Valts Blukis, Arsalan Mousavian, Ankit Goyal, Danfei Xu, Jonathan
  Tremblay, Dieter Fox, Jesse Thomason, and Animesh Garg. 2022.
\newblock \href {https://doi.org/10.48550/arXiv.2209.11302} {Progprompt:
  Generating situated robot task plans using large language models}.
\newblock \emph{CoRR}, abs/2209.11302.

\bibitem[{Song et~al.(2022{\natexlab{a}})Song, Kil, Pan, Sadler, Chao, and
  Su}]{song2022one}
Chan~Hee Song, Jihyung Kil, Tai-Yu Pan, Brian~M Sadler, Wei-Lun Chao, and
  Yu~Su. 2022{\natexlab{a}}.
\newblock One step at a time: Long-horizon vision-and-language navigation with
  milestones.
\newblock In \emph{Proceedings of the IEEE/CVF Conference on Computer Vision
  and Pattern Recognition}, pages 15482--15491.

\bibitem[{Song et~al.(2022{\natexlab{b}})Song, Wu, Washington, Sadler, Chao,
  and Su}]{song2022llm}
Chan~Hee Song, Jiaman Wu, Clayton Washington, Brian~M Sadler, Wei-Lun Chao, and
  Yu~Su. 2022{\natexlab{b}}.
\newblock \href {https://doi.org/10.48550/arXiv.2212.04088} {Llm-planner:
  Few-shot grounded planning for embodied agents with large language models}.
\newblock \emph{CoRR}, abs/2212.04088.

\bibitem[{Su et~al.(2016)Su, Sun, Sadler, Srivatsa, G{\"u}r, Yan, and
  Yan}]{su-etal-2016-generating}
Yu~Su, Huan Sun, Brian Sadler, Mudhakar Srivatsa, Izzeddin G{\"u}r, Zenghui
  Yan, and Xifeng Yan. 2016.
\newblock \href {https://doi.org/10.18653/v1/D16-1054} {On generating
  characteristic-rich question sets for {QA} evaluation}.
\newblock In \emph{Proceedings of the 2016 Conference on Empirical Methods in
  Natural Language Processing}, pages 562--572, Austin, Texas. Association for
  Computational Linguistics.

\bibitem[{Sun et~al.(2020)Sun, Zhang, Cheng, and Qu}]{sun2020sparqa}
Yawei Sun, Lingling Zhang, Gong Cheng, and Yuzhong Qu. 2020.
\newblock \href {https://ojs.aaai.org/index.php/AAAI/article/view/6426}
  {{SPARQA:} skeleton-based semantic parsing for complex questions over
  knowledge bases}.
\newblock In \emph{The Thirty-Fourth {AAAI} Conference on Artificial
  Intelligence, {AAAI} 2020, The Thirty-Second Innovative Applications of
  Artificial Intelligence Conference, {IAAI} 2020, The Tenth {AAAI} Symposium
  on Educational Advances in Artificial Intelligence, {EAAI} 2020, New York,
  NY, USA, February 7-12, 2020}, pages 8952--8959. {AAAI} Press.

\bibitem[{Sutskever et~al.(2014)Sutskever, Vinyals, and Le}]{seq2seq}
Ilya Sutskever, Oriol Vinyals, and Quoc~V. Le. 2014.
\newblock \href
  {https://proceedings.neurips.cc/paper/2014/hash/a14ac55a4f27472c5d894ec1c3c743d2-Abstract.html}
  {Sequence to sequence learning with neural networks}.
\newblock In \emph{Advances in Neural Information Processing Systems 27: Annual
  Conference on Neural Information Processing Systems 2014, December 8-13 2014,
  Montreal, Quebec, Canada}, pages 3104--3112.

\bibitem[{Wang et~al.(2020)Wang, Shin, Liu, Polozov, and
  Richardson}]{wang-etal-2020-rat}
Bailin Wang, Richard Shin, Xiaodong Liu, Oleksandr Polozov, and Matthew
  Richardson. 2020.
\newblock \href {https://doi.org/10.18653/v1/2020.acl-main.677} {{RAT-SQL}:
  Relation-aware schema encoding and linking for text-to-{SQL} parsers}.
\newblock In \emph{Proceedings of the 58th Annual Meeting of the Association
  for Computational Linguistics}, pages 7567--7578, Online. Association for
  Computational Linguistics.

\bibitem[{Wang et~al.(2021)Wang, Wang, Joty, and Hoi}]{wang-etal-2021-codet5}
Yue Wang, Weishi Wang, Shafiq Joty, and Steven~C.H. Hoi. 2021.
\newblock \href {https://doi.org/10.18653/v1/2021.emnlp-main.685} {{C}ode{T}5:
  Identifier-aware unified pre-trained encoder-decoder models for code
  understanding and generation}.
\newblock In \emph{Proceedings of the 2021 Conference on Empirical Methods in
  Natural Language Processing}, pages 8696--8708, Online and Punta Cana,
  Dominican Republic. Association for Computational Linguistics.

\bibitem[{Wei et~al.(2022)Wei, Wang, Schuurmans, Bosma, brian ichter, Xia, Chi,
  Le, and Zhou}]{wei2022chain}
Jason Wei, Xuezhi Wang, Dale Schuurmans, Maarten Bosma, brian ichter, Fei Xia,
  Ed~H. Chi, Quoc~V Le, and Denny Zhou. 2022.
\newblock \href {https://openreview.net/forum?id=_VjQlMeSB_J} {Chain of thought
  prompting elicits reasoning in large language models}.
\newblock In \emph{Advances in Neural Information Processing Systems}.

\bibitem[{Williams and Zipser(1989)}]{williams1989learning}
Ronald~J. Williams and David Zipser. 1989.
\newblock \href {https://doi.org/10.1162/neco.1989.1.2.270} {A learning
  algorithm for continually running fully recurrent neural networks}.
\newblock \emph{Neural Comput.}, 1(2):270--280.

\bibitem[{Xie et~al.(2022)Xie, Wu, Shi, Zhong, Scholak, Yasunaga, Wu, Zhong,
  Yin, Wang, Zhong, Wang, Li, Boyle, Ni, Yao, Radev, Xiong, Kong, Zhang, Smith,
  Zettlemoyer, and Yu}]{xie2022unifiedskg}
Tianbao Xie, Chen~Henry Wu, Peng Shi, Ruiqi Zhong, Torsten Scholak, Michihiro
  Yasunaga, Chien{-}Sheng Wu, Ming Zhong, Pengcheng Yin, Sida~I. Wang, Victor
  Zhong, Bailin Wang, Chengzu Li, Connor Boyle, Ansong Ni, Ziyu Yao,
  Dragomir~R. Radev, Caiming Xiong, Lingpeng Kong, Rui Zhang, Noah~A. Smith,
  Luke Zettlemoyer, and Tao Yu. 2022.
\newblock \href {http://arxiv.org/abs/2201.05966} {Unifiedskg: Unifying and
  multi-tasking structured knowledge grounding with text-to-text language
  models}.
\newblock \emph{CoRR}, abs/2201.05966.

\bibitem[{Ye et~al.(2022)Ye, Yavuz, Hashimoto, Zhou, and
  Xiong}]{ye-etal-2022-rng}
Xi~Ye, Semih Yavuz, Kazuma Hashimoto, Yingbo Zhou, and Caiming Xiong. 2022.
\newblock \href {https://doi.org/10.18653/v1/2022.acl-long.417} {{RNG}-{KBQA}:
  Generation augmented iterative ranking for knowledge base question
  answering}.
\newblock In \emph{Proceedings of the 60th Annual Meeting of the Association
  for Computational Linguistics (Volume 1: Long Papers)}, pages 6032--6043,
  Dublin, Ireland. Association for Computational Linguistics.

\bibitem[{Yih et~al.(2015)Yih, Chang, He, and Gao}]{yih-etal-2015-semantic}
Wen-tau Yih, Ming-Wei Chang, Xiaodong He, and Jianfeng Gao. 2015.
\newblock \href {https://doi.org/10.3115/v1/P15-1128} {Semantic parsing via
  staged query graph generation: Question answering with knowledge base}.
\newblock In \emph{Proceedings of the 53rd Annual Meeting of the Association
  for Computational Linguistics and the 7th International Joint Conference on
  Natural Language Processing (Volume 1: Long Papers)}, pages 1321--1331,
  Beijing, China. Association for Computational Linguistics.

\bibitem[{Yih et~al.(2016)Yih, Richardson, Meek, Chang, and
  Suh}]{yih-etal-2016-value}
Wen-tau Yih, Matthew Richardson, Chris Meek, Ming-Wei Chang, and Jina Suh.
  2016.
\newblock \href {https://doi.org/10.18653/v1/P16-2033} {The value of semantic
  parse labeling for knowledge base question answering}.
\newblock In \emph{Proceedings of the 54th Annual Meeting of the Association
  for Computational Linguistics (Volume 2: Short Papers)}, pages 201--206,
  Berlin, Germany. Association for Computational Linguistics.

\bibitem[{Yu et~al.(2022)Yu, Zhang, Ng, Zhu, Li, Wang, Hu, Wang, Wang, and
  Xiang}]{yu2022decaf}
Donghan Yu, Sheng Zhang, Patrick Ng, Henghui Zhu, Alexander~Hanbo Li, Jun Wang,
  Yiqun Hu, William Wang, Zhiguo Wang, and Bing Xiang. 2022.
\newblock \href {https://doi.org/10.48550/arXiv.2210.00063} {Decaf: Joint
  decoding of answers and logical forms for question answering over knowledge
  bases}.
\newblock \emph{CoRR}, abs/2210.00063.

\bibitem[{Yu et~al.(2021)Yu, Wu, Lin, Wang, Tan, Yang, Radev, Socher, and
  Xiong}]{yu2021grappa}
Tao Yu, Chien{-}Sheng Wu, Xi~Victoria Lin, Bailin Wang, Yi~Chern Tan, Xinyi
  Yang, Dragomir~R. Radev, Richard Socher, and Caiming Xiong. 2021.
\newblock \href {https://openreview.net/forum?id=kyaIeYj4zZ} {Grappa:
  Grammar-augmented pre-training for table semantic parsing}.
\newblock In \emph{9th International Conference on Learning Representations,
  {ICLR} 2021, Virtual Event, Austria, May 3-7, 2021}. OpenReview.net.

\bibitem[{Yu et~al.(2018)Yu, Zhang, Yang, Yasunaga, Wang, Li, Ma, Li, Yao,
  Roman, Zhang, and Radev}]{yu-etal-2018-spider}
Tao Yu, Rui Zhang, Kai Yang, Michihiro Yasunaga, Dongxu Wang, Zifan Li, James
  Ma, Irene Li, Qingning Yao, Shanelle Roman, Zilin Zhang, and Dragomir Radev.
  2018.
\newblock \href {https://doi.org/10.18653/v1/D18-1425} {{S}pider: A large-scale
  human-labeled dataset for complex and cross-domain semantic parsing and
  text-to-{SQL} task}.
\newblock In \emph{Proceedings of the 2018 Conference on Empirical Methods in
  Natural Language Processing}, pages 3911--3921, Brussels, Belgium.
  Association for Computational Linguistics.

\bibitem[{Zhuang et~al.(2022)Zhuang, Qin, Jagerman, Hui, Ma, Lu, Ni, Wang, and
  Bendersky}]{zhuang2022rankt5}
Honglei Zhuang, Zhen Qin, Rolf Jagerman, Kai Hui, Ji~Ma, Jing Lu, Jianmo Ni,
  Xuanhui Wang, and Michael Bendersky. 2022.
\newblock \href {https://doi.org/10.48550/arXiv.2210.10634} {Rankt5:
  Fine-tuning {T5} for text ranking with ranking losses}.
\newblock \emph{CoRR}, abs/2210.10634.

\end{thebibliography}
\bibliographystyle{acl_natbib}

\clearpage
\appendix

\setcounter{table}{0}
\renewcommand\thetable{\Alph{section}.\arabic{table}}
\setcounter{figure}{0}
\renewcommand\thefigure{\Alph{section}.\arabic{figure}}

\section*{Appendices}
In this supplementary material, we provide omitted details in the main text.
\begin{itemize}[topsep=2pt,itemsep=2pt,partopsep=2pt, parsep=2pt]
\item \autoref{appendix:others}: Possible Implementation for Other Tasks
\item \autoref{appendix:candidate}: Candidate Enumeration
\item \autoref{appendix:setup}: Experimental Setup
\item \autoref{appendix:decomposition}: Decomposition by Question Complexity
\item \autoref{appendix:error}: Error Analysis
\item \autoref{appendix:prompts}: Examples of Prompts
\end{itemize}

\section{Possible Implementation for Other Tasks}\label{appendix:others}
In this paper, we choose KBQA as a representative testbed to instantiate \OurMethod without loss of generality. It is worth noting that though~\autoref{alg:highlevel} describes a generic framework for grounded language understanding, the concrete implementation of~\autoref{alg:highlevel} may vary for different tasks. In this section, we shed light on the possible implementation of several other tasks of different nature.
\subsection{Text-to-SQL Parsing}
Similar to KBQA, Text-to-SQL parsing also aims to map a natural language utterance onto a program that can be executed over a relational database (instead of a KB). 
We can define $P_0$ as the set of cell values mentioned in the utterance (similar to entities in KBQA), which should be straightforward to identify (\eg, with string matching).
Also, one necessary step is to convert a SQL query into an algebra tree~\cite{codd1970relational}, similar to what is done by~\citet{rubin-berant-2021-smbop}. 
In this way, the agent can more easily enumerate the candidate programs in a bottom-up manner, which resembles candidate enumeration in KBQA.  
The termination check for text-to-SQL parsing can also be implemented similarly. 
One difference in text-to-SQL is that the search space for schema items is much smaller, so it is possible to enhance \OurMethod with input augmentation, particularly for LLMs. Describing the target database in the prompt (\eg, we can include the schema description of a relational database to Codex) has been proven to be useful for text-to-SQL parsing~\cite{cheng2022binding}. 

\subsection{Interacting with Real-world Environments}
Pangu can also be used for guiding bots that interact with real-world environments, being online websites~\cite{Gur2022UnderstandingHW, Nakano2021WebGPTBQ}, or physical environments~\cite{shridhar2020alfred}. 
Given a complex task to be accomplished in the environment, an agent may decompose it into a sequence of subplans (\eg, making a cup of coffee entails first finding a cup then picking up the cup, etc.; \citet{song2022llm}), and combine it with all executable actions in the environment to enumerate the candidate plans and select the best action with an LM. The termination check could also be implemented easily, where the bot may check the environment state and verify if the task has been accomplished (\eg, whether a cup of coffee has been successfully made). One difference in these cases is that real-world environments often contain information from multiple modalities, thus requiring multi-modal language models~\cite{visualbert, vilbert} that are capable of jointly handling textual, visual, and other modalities.

\section{Candidate Enumeration}\label{appendix:candidate}
Our candidate enumeration for KBQA strictly follows the definition of functions in~\autoref{tab:operators}. Specifically, given a set of current plans $P_t$, to construct the candidate set $C_{t+1}$, for each plan $p_i$ in $P_t$, the agent executes it and gets types and relations that are reachable from the denotation of the plan. For each type $t$, the agent enumerates \fl{(AND t $p_i$)} as a candidate. For each relation $r$, the agent enumerates \fl{(JOIN r $p_i$)} as a candidate. If the denotation of $p_i$ is a numerical value, then four similar candidates with comparatives are also included \fl{(LT/LE/GT/GE r $p_i$)}. In addition, candidate plans with superlatives can be enumerated as \fl{(ARGMAX/ARGMIN $p_i$ r)}. Also, \fl{(COUNT $p_i$)} can always be included to $C_{t+1}$. After checking each $p_i$ independently, the agent then checks each pair of plans $p_i$ and $p_j$ from $P_t$, if the execution of $p_i$ and $p_j$ has an overlap, then \fl{(AND $p_i$ $p_j$)} is also included as a candidate plan. The candidate enumeration process is totally transparent to the LM and can be easily controlled based on different needs.

\begin{table}[t]
    \centering
    \small
    \resizebox{1.0\linewidth}{!}{\begin{tabular}{cll}
    \toprule
     \textbf{Composition Rule}  & \textbf{Signature} & \textbf{Comments} \\
     \midrule
     \fl{JOIN} & $R\times (E\cup E')\rightarrow E'$ & a single hop along an edge\\
     \fl{AND} & $(T\cup E')\times E'\rightarrow E'$ & intersection of two sets\\
     \fl{ARGMAX/ARGMIN} & $(T\cup E')\times R\rightarrow E'$ & superlative aggregations \\
     \fl{LT/LE/GT/GE} & $R\times E\rightarrow E'$&  $</\leq/>/\geq$\\
     \fl{COUNT} & $E'\rightarrow N$& set cardinality\\
    \bottomrule    
    \end{tabular}}
    \caption{Functions in KBQA. We follow the definitions in~\cite{gu-su-2022-arcaneqa}. $R$: relation, $T$: type, $E$: entity, $E'$: a set of entities, $N$: integer.}
    \label{tab:operators}
\end{table}

\section{Experimental Setup}\label{appendix:setup}
\begin{table*}[!th]
    \small
    \centering
    \resizebox{.95\textwidth}{!}{
    \begin{tabular}{lccc}
    \toprule
        \textbf{Model} & \textbf{LMs} &\textbf{Grounding Strategy} &\textbf{Guarantees}\\
        \midrule
       \textbf{ArcaneQA}~\cite{gu-su-2022-arcaneqa} & BERT-base & Constrained Decoding & Grammatical+Faithful\\
       \textbf{RnG-KBQA}~\cite{ye-etal-2022-rng} & BERT-base + T5-base & Input Augmentation & N/A\\
       \textbf{TIARA}~\cite{shu2022tiara} & BERT-base + T5-base & Input Augmentation + Constrained Decoding & Grammatical\\
       \textbf{DecAF}~\cite{yu2022decaf} & DPR + FiD-3B & Input Augmentation & N/A\\
       \textbf{UnifiedSKG}~\cite{xie2022unifiedskg} & T5-base(/large/3B) & Input Augmentation & N/A \\
    \bottomrule
    \end{tabular}}
    \caption{A brief summary of main baseline models.}
    \label{table:baselines}
\end{table*}

\begin{table}[!th]
    \small
    \centering
    \resizebox{.4\textwidth}{!}{
    \begin{tabular}{cccc}
    \toprule
        \textbf{Dataset} & \textbf{Training} &\textbf{Dev} &\textbf{Test}\\
        \midrule
       \GrailQ & 44,337 & 6,763 & 13,231\\
       \GraphQ & 2,381 & $-$ & 2,395\\
       \WebQSP & 3,098 & $-$ & 1,639\\
    \bottomrule
    \end{tabular}}
    \caption{Statistics of KBQA datasets.}
    \label{table:datasets}
\end{table}
\subsection{Datasets Statistics}\label{appendix:datasets}

All three datasets provide gold program annotations. For consistency, we use the converted S-expressions representation provided by~\citet{gu-su-2022-arcaneqa} in our experiments. Concrete statistics of different datasets are shown in~\autoref{table:datasets}.

\subsection{More Details on Baselines}\label{appendix:baselines}
Different LMs and decoding strategies are used in the baseline models.

\noindent \textbf{ArcaneQA}~\cite{gu-su-2022-arcaneqa} is an encoder-decoder model built on top of a BERT encoder. It leverages constrained decoding and incrementally synthesizes a sequence of subprograms, where the constraints come from both the grammar and the execution of existing subprograms, to enforce grammaticality and faithfulness.

\noindent \textbf{TIARA}~\cite{shu2022tiara} first uses BERT to retrieve a set of schema items, which are further used as the input, together with the question, to T5 for plan generation. They also apply constrained decoding but only for grammaticality.

\noindent \textbf{DecAF}~\cite{yu2022decaf} similarly retrieves a relevant subgraph from the KB using DPR~\cite{karpukhin-etal-2020-dense}, and then input the retrieved items to FiD~\cite{izacard-grave-2021-leveraging}, a T5 model fine-tuned for question answering. 

\noindent \textbf{RnG-KBQA}~\cite{ye-etal-2022-rng} first uses BERT to rank a set of enumerated candidate programs (up to a limited complexity), and then uses T5 to edit the top programs into more complex programs. 

\noindent \textbf{UnifiedSKG}~\cite{xie2022unifiedskg} also retrieves a subgraph from the KB as input to T5. The setting of UnifiedSKG is different from other baselines. It assumes the gold schema items are always included in the retrieved subgraph and restricts the number of negative schema items in the subgraph (\ie, at most 20 schema items for \GrailQ). 
It is thus a less fair comparison for other methods, but we include it anyway because it is a representative way of autoregressive plan generation using a large LM.

A summary of the baselines can be found in~\autoref{table:baselines}. 

\subsection{Implementation Details}\label{appendix:implementation}
For \GrailQ we use the entity linking results from TIARA. 
For \WebQSP, we get that from ELQ~\cite{li-etal-2020-efficient}, which is also used by our baseline models. 
For \GraphQ, get that from ArcaneQA. 
The entity proposals for the input utterance form the initial plans ($P_0$) for our search process. 
We use beam size 5 for all of our fine-tuning experiments. 
We run our experiments with T5-3B using a single NVIDIA A100 80GB card, while for all other fine-tuning experiments, we run them using 4$\times$ NVIDIA A6000 48GB cards.

For our experiments with Codex, we use a beam size of 2 and a max number of candidates of 1,000 for speed concerns, which to some extent sacrifices the performance. 
As the first endeavor towards enabling few-shot KBQA with LLMs, we did not tune the hyper-parameters very hard. 
The only thing we tuned is the scoring function. 
We tune the scoring function using 10-shot training data from \GrailQ with cross-validation. 
If we directly use $P(c|u)$ as our scoring function $s(u, c)$ in~\autoref{section:lm-evaluation}, Codex tends to favor programs with repeated relations. 
As a result, we add a penalizing factor to $P(c|u)$, and define $s(u, c)$ as $P(c|u)\times \eta^n$, where $\eta\in [0, 1]$ is a hyper-parameter, and $n$ is the maximal occurrences of a relation in a program. We set $\eta=0.7$ based on cross-validation using the 10 training examples.

Finally, a small percentage of questions (around 5\%) in \GraphQ and \GrailQ do not have a topic entity (\eg, \textit{``who is the heaviest film director?"} from \GrailQ, whose target program is \fl{(ARGMAX film.director people.person.weight\_kg)}). For these questions, we use the answer types (\eg, \fl{film.director}) predicted in~\citet{gu-su-2022-arcaneqa} as our initial state $P_0$.

\section{Decomposition by Question Complexity}\label{appendix:decomposition}
\begin{table}[!h]
  \begin{subtable}{\linewidth}
  \small
    \centering
    \resizebox{.9\textwidth}{!}{\begin{tabular}{lcccc}
    \toprule
     \textbf{\# of relations} & \textbf{1} & \textbf{2} & \textbf{3} & \textbf{4}\\\midrule
    RnG-KBQA & 79.2 & 74.8& 44.4 & \textbf{100.0} \\
    ArcaneQA & 80.9 & 71.1& 37.7& \textbf{100.0}\\
    TIARA & 85.6 & 75.8 & \textbf{48.5} & 83.3 \\
    \OurMethod w/ T5-3B & \textbf{87.0} & \textbf{78.4} & 48.1 & 83.3 \\
    \OurMethod w/ Codex (100-shot) & 73.9 & 43.4 & 33.0 & 16.7 \\
    \bottomrule
    \end{tabular}}
        \caption{\GrailQ}
\label{table:fine_grail}
\end{subtable}
  \begin{subtable}{\linewidth}
  \small
\centering
    \resizebox{0.8\textwidth}{!}{\begin{tabular}{lccc}
    \toprule
     \textbf{\# of relations} & \textbf{1} & \textbf{2} & \textbf{3} \\\midrule
    ArcaneQA & 48.2 & 19.3 & 9.6\\
    \OurMethod w/ T5-3B & \textbf{72.3} & \textbf{55.5} & \textbf{27.8} \\
    \OurMethod w/ Codex (100-shot) & 52.2 & 36.1 & 17.5\\
    \bottomrule
    \end{tabular}}
        \caption{\GraphQ}
\label{table:fine_gq}
\end{subtable}
\caption{F1 decomposition by program complexity on \GrailQ's dev set and \GraphQ's test set.}
\label{table:fine}
\end{table}
We present a fine-grained analysis of \OurMethod with T5-3B and Codex (100-shot) on questions of different complexity, measured by the number of relations in the gold program, in~\autoref{table:fine}. 
For \GrailQ, we report the performance on its dev set because the test set is hidden. 
\OurMethod performs competitively across all complexity. 
Note that there are only two questions in \GrailQ's dev set with 4 relations, so the results on that may not be indicative. 
On \GraphQ, \OurMethod significantly outperforms ArcaneQA. 
The F1 of \OurMethod with T5-3B is almost three times higher than ArcaneQA on questions with 2 and 3 relations. 
Interestingly, \OurMethod with Codex also outperforms ArcaneQA considerably on questions with 2 and 3 relations. 
These findings suggest the superiority of \OurMethod in generalizing to more complex programs\nop{, but also leave room for further improvement, especially for in-context learning}.

\section{Error Analysis}\label{appendix:error}
We analyze 200 incorrect predictions (\ie, EM=0) randomly sampled from \GrailQ's dev set for our best model (\ie, T5-3B). The major errors are due to unidentified topic entities during entity linking (62\%).\footnote{The recall of entity linking on \GrailQ is 88.6\%~\cite{shu2022tiara}}
Also, \OurMethod tends to include unrelated entities provided by the entity linker into the final programs (6.5\% of the errors), this is because \OurMethod is fine-tuned with gold entities only, and thus does not learn to handle unrelated entities. In addition, wrong termination check corresponds to 12.5\% of the errors, indicating a venue for better enforcing the partial order to \OurMethod. Apart from these errors, 10.5\% of the mistakes are due to ambiguous annotations or annotation errors in \GrailQ. The remaining error types include wrong comparators, answer types, and relations (particularly relations involve a subtle direction like \fl{cvg.computer\_game\_engine.predecessor\_engine}).

In addition, for in-context learning with Codex (100-shot), we also randomly sample 200 wrong predictions from \GrailQ's dev set. In addition to 22\% errors caused by missing entities, the most common errors (25.5\%) are due to wrong schema items. Distinguishing gold schema items from confusing ones is challenging for in-context learning. Also, missing constraints (16.5\%) and missing relations (10\%) are another two major error types, because we use a small batch size (\ie, 2) for Codex and the model tends to prefer short programs. These two error types are also related to wrong termination check. Finally, there are 12\% wrong functions. The error types of \OurMethod w/ Codex are very different from \OurMethod w/ T5-3B. This is because for a complex task like KBQA, the performance of in-context learning with \OurMethod still largely lags behind fine-tuning. Particularly, fine-tuning methods directly learn the partial order among programs during training, while Codex needs to implicitly infer a partial order by itself, which is not directly shown in the demonstrations.  As a result, \OurMethod w/ Codex makes more trivial mistakes that fine-tuning methods can easily avoid. More advanced in-context learning techniques to close this gap remains to be explored.

\section{Examples of Prompts}\label{appendix:prompts}
\begin{figure*}[!h]
\centering
\includegraphics[width=\linewidth]{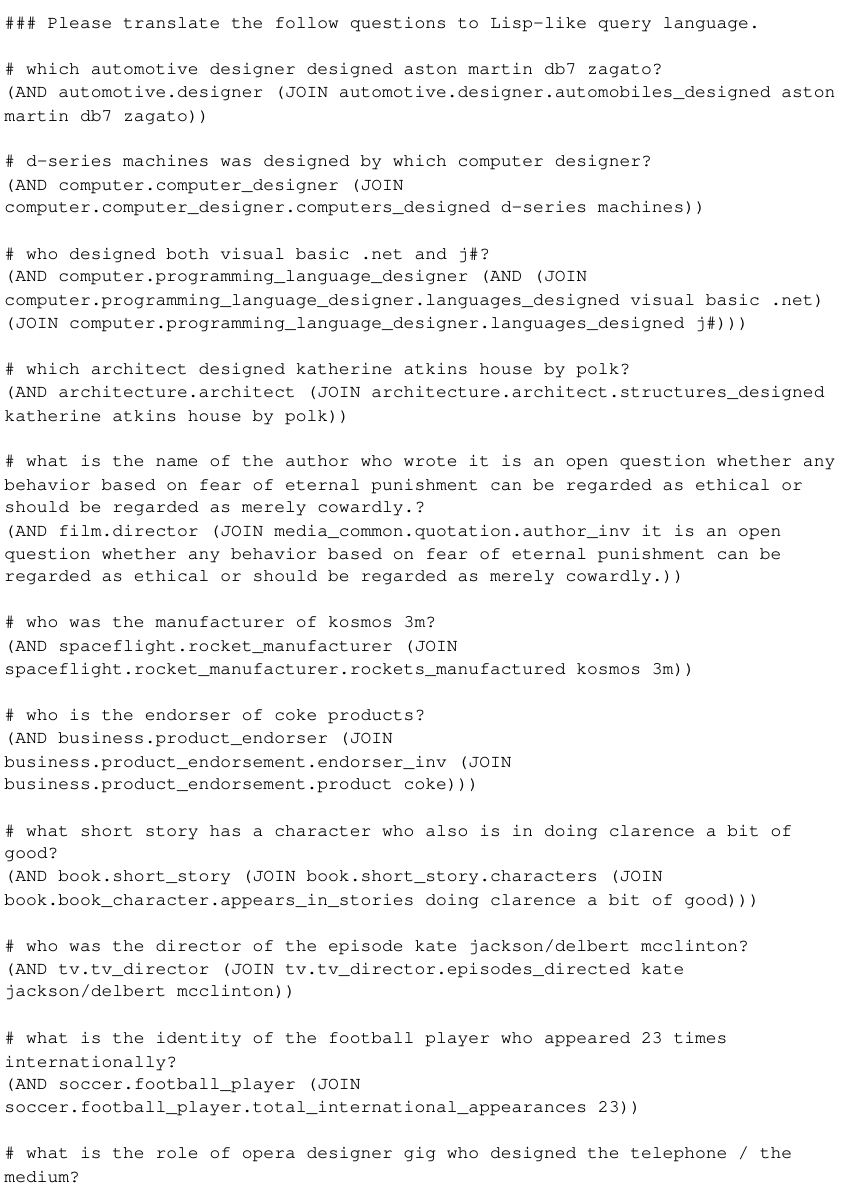}
\caption{Example Prompt (i) for question \textit{``what is the role of opera designer gig who designed the telephone / the medium?"}.}
\label{fig:prompt}
\vspace{-10pt}
\end{figure*}

\begin{figure*}[!h]
\centering
\includegraphics[width=\linewidth]{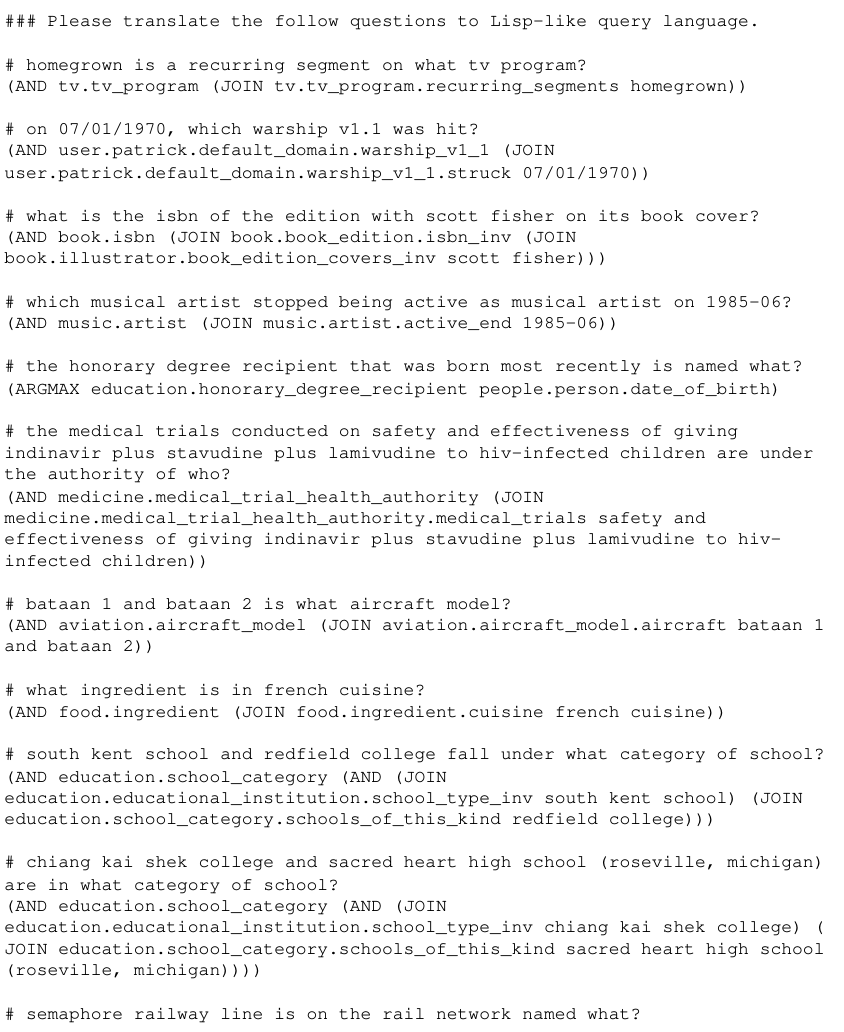}
\caption{Example Prompt (ii) for question \textit{``semaphore railway line is on the rail network named what?"}.}
\label{fig:prompt2}
\vspace{-10pt}
\end{figure*}

We show two examples of prompts with 10 in-context samples retrieved from the 100 training data pool in~\autoref{fig:prompt} and~\autoref{fig:prompt2} for two different questions from \GrailQ's dev set. Our prompt design is very straightforward. More advanced prompting techniques for \OurMethod remains to be explored.

\end{document}